\documentclass{article}

\usepackage{microtype}
\usepackage{graphicx}
\usepackage{subfigure}
\usepackage{booktabs} 

\usepackage{hyperref}
\usepackage{xurl}
\usepackage{multirow}
\usepackage[table]{xcolor} 
\usepackage{tabularx}
\usepackage{array}
\usepackage{amsmath}
\usepackage{xcolor}
\usepackage{pifont}
\usepackage{xspace}
\usepackage{enumitem}
\usepackage{fontawesome5}

\makeatletter
\newcommand{\ie}{\emph{i.e.}\@ifnextchar.{\!\@gobble}{}}
\newcommand{\eg}{\emph{e.g.}\@ifnextchar.{\!\@gobble}{}}
\newcommand{\etc}{etc\@ifnextchar.{}{.\@}}
\makeatother

\definecolor{ansblue}{HTML}{2F4FB3}
\newcommand{\ans}[1]{$\rightarrow$~\textcolor{ansblue}{\textbf{#1}}}
\newcolumntype{Y}{>{\raggedright\arraybackslash}X}
\newcommand{\bench}{\textsc{ScreenMemEval}\xspace}
\newcommand{\benchshort}{\textsc{SME}\xspace}
\newcommand{\benchA}{\mbox{SME-H1}\xspace}   
\newcommand{\benchB}{\mbox{SME-H2}\xspace}   
\newcommand{\benchC}{\mbox{SME-Agent}\xspace} 
\newcommand{\method}{\textsc{StreamEmbed}\xspace}

\definecolor{clBlue}{HTML}{30B0F0}\definecolor{clPink}{HTML}{F07080}
\colorlet{accA}{clBlue!3}\colorlet{accB}{clBlue!8}\colorlet{accC}{clBlue!15}\colorlet{accD}{clBlue!24}\colorlet{accE}{clBlue!34}\colorlet{accF}{clBlue!45}
\colorlet{latA}{clPink!3}\colorlet{latB}{clPink!8}\colorlet{latC}{clPink!15}\colorlet{latD}{clPink!24}\colorlet{latE}{clPink!34}\colorlet{latF}{clPink!45}

\usepackage[accepted]{mlsys2025}
\makeatletter
\renewcommand{\printAffiliationsAndNotice}[1]{%
\stepcounter{@affiliationcounter}%
{\let\thefootnote\relax\footnotetext{\hspace*{-\footnotesep}\ifdefined\isaccepted #1\fi%
\forloop{@affilnum}{1}{\value{@affilnum} < \value{@affiliationcounter}}{
\textsuperscript{\arabic{@affilnum}}\ifcsname @affilname\the@affilnum\endcsname%
\csname @affilname\the@affilnum\endcsname%
\else
{\bf AUTHORERR: Missing \textbackslash{}mlsysaffiliation.}
\fi
}.
}
}
}
\makeatother

\mlsystitlerunning{\method: Real-Time Indexing from Pixels for Screenshot Retrieval}
\begin{document}

\twocolumn[
\mlsystitle{\texorpdfstring{
\vspace{-5mm} 
\raisebox{-0.32\height}
{\includegraphics[height=2.1em,trim=35 110 0 115,clip]{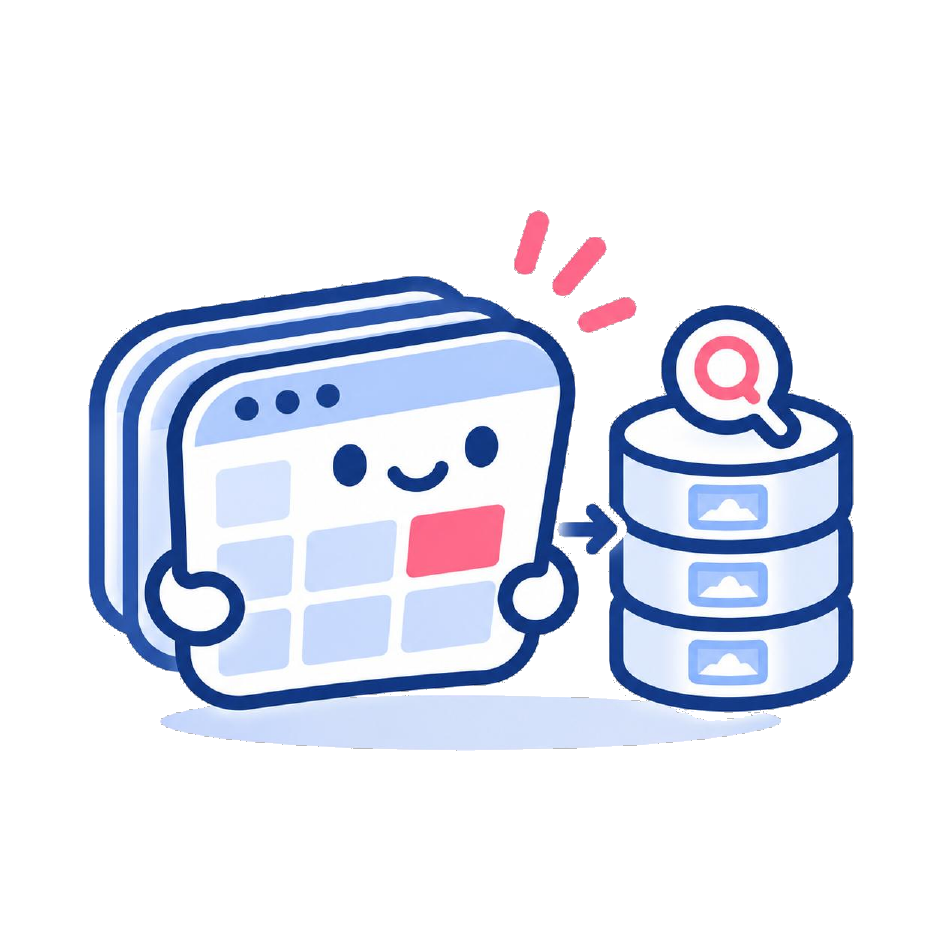}}\hspace{0.25em}}{}\method: Real-Time Indexing from Pixels \\for Screenshot Retrieval}



\mlsyssetsymbol{equal}{*}
\mlsyssetsymbol{corr}{$\dagger$}

\begin{mlsysauthorlist}
\mlsysauthor{Dayoon Ko}{berkeley,snu}
\mlsysauthor{Yichuan Wang}{berkeley}
\mlsysauthor{Soyeong Jeong}{kaist}
\mlsysauthor{Joseph E.~Gonzalez}{berkeley}
\mlsysauthor{Gunhee Kim}{snu}
\mlsysauthor{Sewon Min}{berkeley,corr}
\end{mlsysauthorlist}


\vskip 0.1in
{\centering\small
\faGithub~\href{https://github.com/StarTrail-org/StreamEmbed}{github.com/StarTrail-org/StreamEmbed}
\\
\raisebox{-0.8em}
{\includegraphics[height=2em]{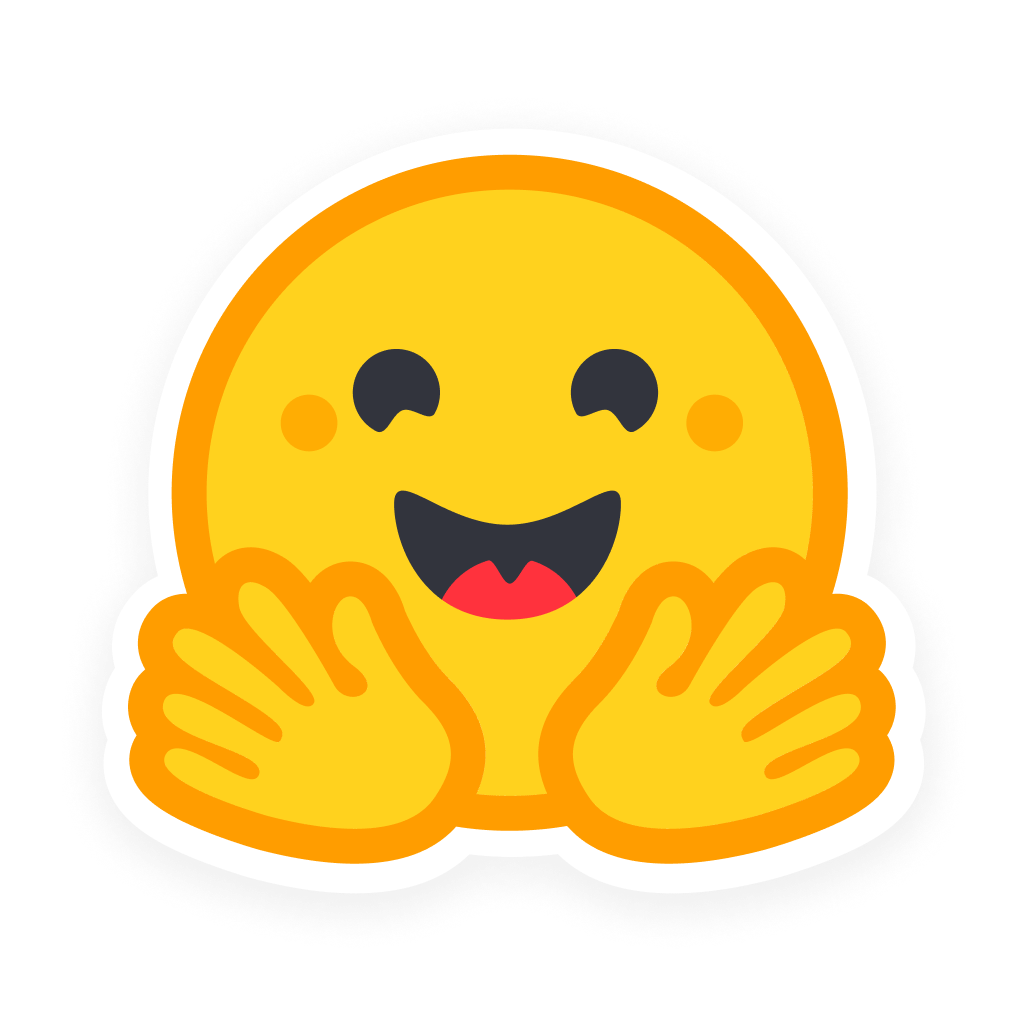}}~\href{https://huggingface.co/datasets/dayoon/ScreenMemEval}{huggingface.co/datasets/dayoon/ScreenMemEval}
\par
\vspace{-2mm}}

\mlsysaffiliation{berkeley}{University of California, Berkeley}
\mlsysaffiliation{snu}{Seoul National University}
\mlsysaffiliation{kaist}{KAIST}

\mlsyskeywords{Machine Learning, MLSys}

\vskip 0.3in

\begin{abstract}
Screen history can serve as searchable visual memory for people and agents, but unlike an offline corpus, it must be encoded and indexed as each screenshot arrives. OCR-based indexing is fast but loses visual information; visual encoding takes over two seconds. Yet screens change incrementally: most pixels repeat within and across screenshots. This raises a question: \emph{do visual encoders need to recompute their representations from scratch for every new screenshot?} We present \method, a system for incremental visual embedding that exploits this redundancy. Guided solely by pixel differences, it combines \emph{spatial token pruning}, which skips blocks that repeat a neighbor, with \emph{temporal KV reuse}, which reuses cached keys and values for blocks unchanged across screenshots. A small learned adapter compensates for pruning and reuse to preserve fidelity to full-screenshot embeddings. On \bench, our benchmark of 10,426 questions over 184 hours of screen histories from two humans and a computer-use agent, \method sets a new accuracy--latency Pareto frontier on an M2 Pro GPU: it encodes up to 3.7$\times$ faster than full-screenshot encoding without sacrificing accuracy and improves accuracy over OCR-based indexing by up to 6.3 Recall@10 points at matched latency.
A memory-efficient vector index stores a simulated year of screenshots in 0.13\,GB of RAM. Together, these results make real-time indexing of screen history in its native visual form practical, laying a foundation for fully local visual assistants.
\end{abstract}
]

\printAffiliationsAndNotice{\textsuperscript{$\dagger$}Corresponding author.}  

\section{Introduction}

\begin{figure*}[t!]
\centering
\includegraphics[width=\textwidth]{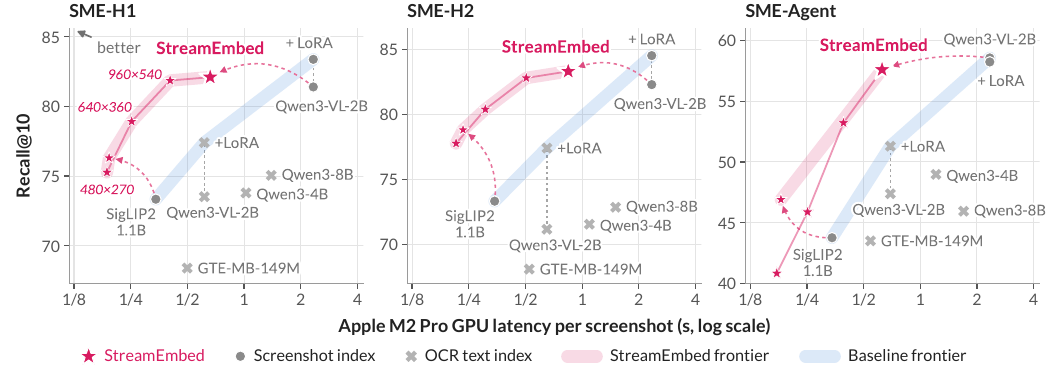}
\vspace{-2mm}
\caption{\textbf{\method moves the accuracy--latency frontier.}
Recall@10 against indexing latency per screenshot (log scale) on an M2 Pro GPU, on the three histories of \bench. \method runs on Qwen3-VL-Embedding-2B at several input resolutions and on SigLIP~2.
\method and the +LoRA baselines are fine-tuned on the training split of \benchA, and all results are on held-out test sets (\S\ref{sec:bench-labels}).
We run baselines as released, while \method uses a linear patch embedding for Qwen-VL (\S\ref{sec:eval-setup}).
}
\label{fig:intro}
\end{figure*}

Much of daily computer work passes through laptop screens: reading papers, browsing products, and messaging colleagues. Yet people remember few of the details and may later ask: ``Which paper had the pink and blue Pareto frontier curves I saw last week?'' A searchable screen history could help people recover these details, and also provide persistent memory for locally running agents~\citep{packer2023memgpt, wu2025longmemeval}. Because screenshots capture every application without specialized integration, they offer a unified interface across text and visual contents~\citep{hu2020screentrack}. Answering questions about past activity then becomes a retrieval problem over a continuously growing screenshot history.

The key challenge is maintaining this visual memory continuously without exhausting system resources. 
Unlike a fixed corpus that is indexed offline~\citep{karpukhin2020dense}, this setting requires \emph{real-time indexing}: encoding and indexing each screenshot as it arrives, ideally locally to keep private content on device, while consuming only a small fraction of the laptop's capacity.

Converting each screenshot into OCR text, as existing open-source tools do~\citep{openrecall, rem}, is fast but loses what only pixels capture~\citep{wang2026pixelrag}.
Direct visual embedding preserves this information but takes 2.3\,s per screenshot on the GPU. At one screenshot every 5\,s, this uses the GPU for 28 minutes per hour of the user's work, and at one every 2\,s, it cannot keep pace. Even commercial systems, such as Microsoft Recall~\citep{microsoftrecall}, require dedicated hardware like NPUs and at least 16\,GB of RAM.
Thus, the central question is: can we retain visual retrieval quality while keeping continuous encoding lightweight enough to run silently in the background?

\begin{figure*}[t!]
\centering
\includegraphics[width=0.98\textwidth]{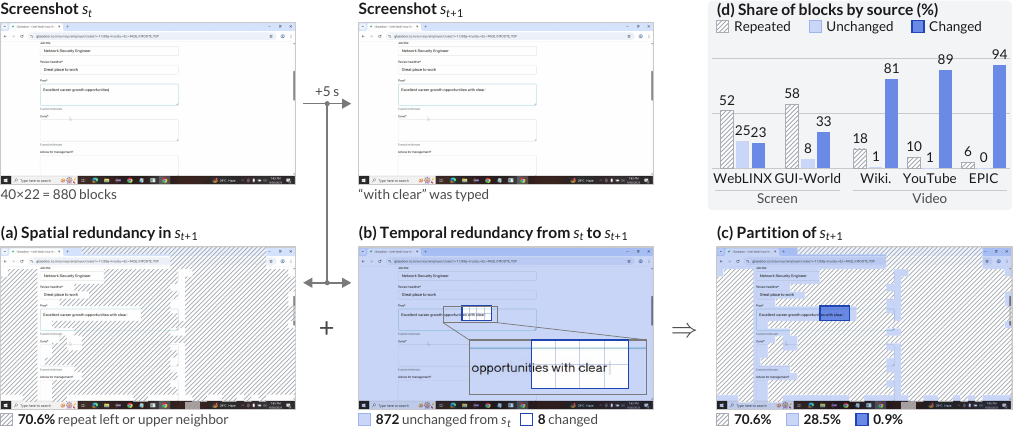}
\vspace{-3mm}
\caption{
\textbf{Screenshots repeat most of their blocks, unlike natural video.} (a) Blocks of $s_{t+1}$ that repeat their left or upper neighbor. (b) Blocks of $s_{t+1}$ unchanged from $s_t$. (c) (a) and (b) together split $s_{t+1}$ into repeated, unchanged, and changed blocks. (d) Share of repeated, unchanged, and changed blocks per source ~\citep{lu2024weblinx, chen2025gui, damen2022rescaling} (Wiki.: Wikimedia Commons).
}
\vspace{-3mm}
\label{fig:redundancy}
\end{figure*}

Screen histories, however, offer a particular opportunity: they change incrementally and are largely redundant (Figure \ref{fig:redundancy}).
Background elements often repeat within and across screenshots, and typing or moving the cursor changes only a few regions, leaving the rest untouched. This raises a question: \emph{do we need to compute representations from scratch for every screenshot?}

We propose \method, which encodes screenshots as a stream rather than as isolated images, exploiting this redundancy. 
Within a screenshot, it drops pixel blocks that merely duplicate a neighboring block (spatial pruning), cutting the computation by 33--41\%. Across screenshots, it recomputes only the changed blocks and reuses cached keys and values for the rest in both the vision encoder and the language model (temporal reuse), cutting the remaining computation by a further 34--57\%.
These two techniques are largely complementary, and combining them further reduces the computation per screenshot by 57--71\%.
A small adapter merged into the weights compensates for pruning and reuse, keeping the resulting embedding close to that of the full screenshot.

Evaluating \method presents another challenge, since no suitable benchmark is available. Such a benchmark needs a screenshot history, questions about it, and the screenshots that answer each question, but existing benchmarks lack a screenshot history~\citep{wu2026longmemeval}, questions about the past~\citep{xie2024osworld}, or relevance labels and public data~\citep{wang2025mirix}. We therefore build \bench (\benchshort), consisting of 10,426 questions drawn from three screenshot histories, each labeled with the screenshots that answer it. Two histories come from human workers on Windows in a public dataset of computer-use recordings~\citep{psai_computer_use_2025}: \benchA spans 117 days with 88 hours of screen time, and \benchB spans 57 days with 18 hours. The third, \benchC, spans 43 days with 78 hours, recorded on Ubuntu from a computer-use agent (the frontier model GPT-6.1 Sol~\citep{openai2026gpt61sol}) performing the tasks of myPCBench~\citep{jang2026mypcbench}.

On \benchshort, \method on Qwen3-VL-Embedding-2B \citep{li2026qwen3} sets a new accuracy--latency Pareto frontier (Figure \ref{fig:intro}). 
Against baselines on the same backbone, fine-tuned on the same data, it is up to 6.3 points more accurate than an OCR-based index at similar latency, and up to 3.7$\times$ faster than a screenshot index while retaining over 98\% of its accuracy.
To cut latency, one can lower the input resolution or use a smaller encoder. When we apply these to both \method and the baselines, \method is more accurate at every latency, by up to 11.4 points (\S\ref{sec:eval-encoders}). We also show that binary codes of the embeddings reduce memory usage without sacrificing accuracy on \benchshort. With them, an index of five simulated years of screenshots fits in under 1\,GB of RAM, and search takes under 0.1\,s. Together, this work makes real-time indexing of screen history in its native visual form practical and efficient, paving the way for fully local, visual computer-use assistants.

In summary, our contributions are as follows:
\begin{itemize}
[leftmargin=1.2em, itemsep=8pt, topsep=2pt, parsep=0pt]
    \item We introduce real-time indexing of screenshot streams: continuously building searchable visual memory on local hardware, where existing methods face a trade-off between lossy OCR text and slow screenshot embeddings.
    \item We propose \method, which exploits redundancy within and across screenshots through spatial token pruning and temporal KV reuse. Both use pixels alone, and a small adapter merged into model weights preserves fidelity to full-screenshot embeddings.
    \item We build \bench, consisting of 10,426 questions over 184 hours of screen histories from two humans and a computer-use agent.
    \item On \bench, \method achieves the new latency-accuracy frontier among evaluated methods: it runs up to 3.7$\times$ faster than the base screenshot encoder at matched accuracy. We also show its index fits five years of screenshots in under 1\,GB of RAM.

\end{itemize}

\section{Problem Setup}
\label{sec:background}

\textbf{Real-time indexing.} While a human or an agent uses a computer, its screen history arrives as a stream of screenshots $\mathcal{S} = (s_1, s_2, \dots)$, one every $\Delta$ seconds. Each screenshot $s_t$ is indexed as it arrives, building a memory for future questions. To keep private content local, indexing runs on the user's device. Here, indexing $s_t$ means encoding it into a vector $\mathbf{v}_t$ with an embedding model $E$ and adding $\mathbf{v}_t$ to a growing index.

Indexing is \emph{real-time} if it keeps pace with the stream, so that screenshots never accumulate faster than they are indexed. Let the indexing latency $L$ be the mean time to index one screenshot, including any preprocessing such as OCR. Keeping pace thus requires, at minimum, $L \le \Delta$; practical use further calls for a small $L$, since the device remains in use for other work during indexing.

Retrieval runs only when a question $q$ arrives: the same model $E$ embeds $q$, and $\mathcal{T}_k(q) = (s_{(1)}, \dots, s_{(k)})$ returns the $k$ indexed screenshots with the highest cosine similarity to $q$, as in dense retrieval~\citep{karpukhin2020dense}. This takes about 0.4\,s (Table~\ref{tab:index-scaling}) and runs once per question rather than every $\Delta$ seconds, so only indexing is time-constrained.

The challenge thus lies in indexing, which has two competing goals: preserving in each $\mathbf{v}_t$ whatever a future question may need, and keeping $L$ small. Accordingly, \S\ref{sec:eval-main} reports retrieval accuracy against $L$ for each method.

\textbf{Questions and evaluation.} Each history in \benchshort consists of task recordings: contiguous spans of screenshots, each covering one task. A question $q$ is answered by $n$ task recordings, and all screenshots of the $m$-th recording ($1 \le m \le n$) form the relevant set $\mathcal{S}^{+}_m(q)$. For evaluation, each history is fully indexed before any question is asked, and retrieval accuracy is the fraction of these recordings that $\mathcal{T}_k(q)$ hits with at least one screenshot,
\begin{equation}
\mathrm{Recall@}k(q) = \frac{1}{n}\,\big|\{\, m : \mathcal{S}^{+}_m(q) \cap \mathcal{T}_k(q) \neq \emptyset \,\}\big|,
\label{eq:recall}
\end{equation}
averaged over test questions~\citep{thakur2021beir}.

\textbf{Text or screenshots.} An index can apply $E$ either to the OCR text of each screenshot $s_t$, as a \emph{text index}, or to $s_t$ itself, as a \emph{screenshot index}. Existing open-source tools build text indexes~\citep{screenpipe, openrecall, rem}, but OCR text often discards much of what the screen shows~\citep{ma2024unifying, faysse2025colpali, wang2026pixelrag}: it mixes the answer with every other string on the screen, such as menu labels and text from other windows; it loses the layout that tells them apart; and it drops state that only the pixels show, such as which option is selected.

\section{Visual Encoding}
\label{sec:redundancy}

A screenshot index encodes each arriving screenshot with a vision-language embedding model. This section describes the challenge, encoding as the main cost of indexing (\S\ref{sec:challenges}), and the opportunity, redundant blocks that account for most of it (\S\ref{sec:redundancy-measure}).

\subsection{Challenges}
\label{sec:challenges}

Indexing a screenshot has two steps: encoding and insertion. Insertion takes under a millisecond, so nearly all of the cost lies in encoding. A vision-language embedding model such as Qwen3-VL-Embedding encodes a screenshot with a vision encoder and passes the resulting hundreds of visual tokens through its language model, whereas a text index passes only the OCR text through the language model.

As Table~\ref{tab:ocr} shows, encoding a screenshot takes 2.33\,s on an M2 Pro GPU: 7.5$\times$ as long as encoding its OCR text, and 3.6$\times$ the total latency of a text index even with OCR included. At one screenshot every $\Delta = 5$\,s, encoding thus occupies the GPU nearly half of the time, on a laptop that remains in use for other work. This cost likely explains why existing tools index OCR text, despite a 10.1-point drop in R@10~\citep{screenpipe, openrecall, rem}. Moreover, it recurs for every screenshot, since screenshot indexes encode each one from scratch.

\vspace{-1mm}
\subsection{Opportunities}
\label{sec:redundancy-measure}
\vspace{-1mm}
Encoding each screenshot from scratch ignores how much of a screen repeats, both within a screenshot and across screenshots. (i) Spatial redundancy arises within a screenshot, where backgrounds, margins, and blank panels span large areas (Figure~\ref{fig:redundancy}(a)). (ii) \emph{Temporal redundancy} arises across screenshots, where typing or moving the cursor changes a few regions and leaves the rest of the screen as it was (Figure~\ref{fig:redundancy}(b)).  Natural video has little of either: texture varies within a frame, and camera motion alters most pixels between frames.

Figure~\ref{fig:redundancy}(d) quantifies this. In screen recordings, 67--77\% of the pixel blocks that become visual tokens repeat a neighbor exactly or are unchanged from the previous screenshot, against 6--19\% in natural video. An encoder that treats every screenshot as new thus spends most of its work on redundant tokens. It could instead skip blocks that repeat a neighbor, reuse the computation of blocks unchanged since the previous screenshot, and recompute only the rest. \method builds on this idea (\S\ref{sec:ditto}).

\begin{table}[t]
\centering
\small
\setlength{\tabcolsep}{4pt}
\renewcommand{\arraystretch}{1.0}
\vspace{-2mm}
\caption{Mean R@10 and per-screenshot latency (s) of Qwen3-VL-Embedding-2B indexes on the \benchshort. Enc.: encoding; Insert: adding the embedding to the index; Total: OCR + Enc. + Insert.}
\vspace{1mm}
\label{tab:ocr}
\begin{tabular*}{\columnwidth}{@{\extracolsep{\fill}}lccccc@{}}
\toprule
 & & \multicolumn{4}{c}{Latency (s)} \\
\cmidrule(l){3-6}
Index & R@10 & OCR & Enc. & Insert & Total \\
\midrule
Text & 64.0 & 0.34 & 0.31 & $<$0.001 & \textbf{0.65} \\
Screenshot & \textbf{74.1} & -- & 2.33 & $<$0.001 & 2.33 \\
\bottomrule
\end{tabular*}
\vspace{-6mm}
\end{table}

\vspace{-1mm}
\section{\method} 
\label{sec:ditto}

\begin{figure*}[t!]
\centering
\includegraphics[width=\textwidth]{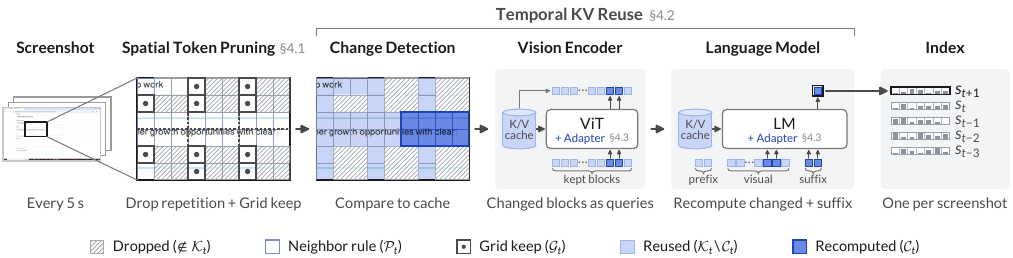}
\vspace{-5mm}
\caption{\textbf{Indexing with \method.} A screenshot arrives every $\Delta = 5$\,s. Spatial pruning drops blocks that repeat their left or upper neighbors (\S\ref{sec:prune}), and change detection flags kept blocks that are new or have changed beyond a threshold (\S\ref{sec:reuse}). The ViT and LM recompute only the flagged blocks (and the suffix), reusing cached keys and values for the rest. The last token's final hidden state is the embedding of the screenshot.}
\vspace{-3mm}
\label{fig:pipeline}
\end{figure*}

\vspace{-1mm}
\method exploits this redundancy (\S\ref{sec:redundancy-measure}) by encoding screenshots as a stream, computing only what has not been shown before. Spatial pruning removes redundancy within a screenshot (\S\ref{sec:prune}), temporal reuse removes redundancy across screenshots by recomputing only changed blocks (\S\ref{sec:reuse}), and a consistency adapter keeps the embedding of the reduced input close to that of the full screenshot (\S\ref{sec:adapter}). Figure~\ref{fig:pipeline} shows how a screenshot is indexed.

\vspace{-1mm}
\paragraph{Embedding a screenshot.} A vision-language embedding model~\citep{ma2024unifying}, such as Qwen3-VL-2B~\citep{li2026qwen3}, encodes a screenshot in three steps. First, a vision encoder (ViT)~\citep{dosovitskiy2021image} turns each $16{\times}16$ patch into a vector. Second, each $2{\times}2$ group of patches, a $32{\times}32$ pixel block, forms one visual token. Third, a language model (LM) reads these tokens between a fixed text prefix and suffix, and its normalized last-layer hidden state at the last token is the embedding $E(s)$. The ViT and the LM stack self-attention layers, in which each token forms a query, key, and value and attends to the keys and values of all tokens, bidirectionally in the ViT and causally in the LM. A $1280{\times}720$ screenshot, resized to $1280{\times}704$, becomes a $22{\times}40$ grid of 880 blocks and tokens.

\vspace{-1mm}
\paragraph{Notation.} We write $b_{t,i}$ for the block at grid position $i$ of $s_t$, so $b_{t,i}$ and $b_{t+1,i}$ cover the same screen region, and $\|b - b'\|_\infty$ for the largest absolute pixel difference between blocks $b$ and $b'$ over all RGB channels.

\subsection{Spatial Token Pruning}
\label{sec:prune}

We prune before the ViT, so a dropped block is processed by neither the ViT nor the LM.

\vspace{-1mm}
\paragraph{Neighbor rule.} We write $\ell(i)$ and $u(i)$ for the left and upper neighbors of block $i$. A block is dropped if it repeats either neighbor within $\tau$, 8 of 255 levels, and kept otherwise,
\begin{equation}
\mathcal{P}_t = \bigl\{\, i \;:\;
\begin{aligned}[t]
&\|b_{t,i}-b_{t,\ell(i)}\|_\infty > \tau \\
&\text{and}\ \|b_{t,i}-b_{t,u(i)}\|_\infty > \tau \,\bigr\}.
\end{aligned}
\label{eq:prune}
\end{equation}
A block on the left or top edge is compared only with the neighbor it has, and the top-left block is always kept. Kept blocks keep their original position embeddings in the ViT and the LM, so each token still encodes where its block lies.

\vspace{-1mm}
\paragraph{Grid keep.} The neighbor rule reduces a large uniform region, such as a blank panel, to a few blocks, which shifts the embedding away from that of the full screenshot. We thus also keep $\mathcal{G}_t$, one block of each uniform region in every fixed $3{\times}3$ tile of blocks, so the kept set is $\mathcal{K}_t = \mathcal{P}_t \cup \mathcal{G}_t$.

\citet{wang2026pixelprune} decide block by block in raster order, so each decision depends on all earlier ones. We instead compare each block with its neighbors in the original screenshot, so all blocks are decided independently and in parallel.

\subsection{Temporal KV Reuse}
\label{sec:reuse}

We reuse the keys and values of earlier screenshots and recompute only changed blocks. Reuse is exact only if every token a reused token attends to is unchanged. Otherwise, it approximates, and periodic refresh bounds the error.

\vspace{-1mm}
\paragraph{Change detection.} Let $\hat{b}_i$ be block $i$ as it was when its cached keys and values were last computed. Among the kept blocks $\mathcal{K}_t$, we recompute those that have no cache or have changed by more than $\delta$ since then,
\begin{equation}
\mathcal{C}_t = \bigl\{\, i \in \mathcal{K}_t \;:\;
\begin{aligned}[t]
&\hat{b}_i \text{ does not exist} \\
&\text{or}\ \|b_{t,i}-\hat{b}_i\|_\infty > \delta \,\bigr\},
\end{aligned}
\label{eq:change}
\end{equation}
with $\delta = \tau$. Comparing with $\hat{b}_i$ rather than $b_{t-1,i}$ keeps every reused block within $\delta$ of $\hat{b}_i$, even when it changes slowly over many screenshots.

\textbf{ViT.} In every layer, we recompute the queries, keys, and values of the blocks in $\mathcal{C}_t$ and update their cache. They attend to all kept blocks, taking the keys and values of unchanged ones from the cache, so each changed block still sees every kept block. The outputs of the other kept blocks come from the cache. The ViT approximates in that an unchanged block does not see what changed elsewhere on the screen.

\textbf{LM.} In the LM, we likewise recompute, in every layer, the visual tokens of $\mathcal{C}_t$ and the fixed suffix and update their cache. They attend causally to all earlier tokens, taking the keys and values of the others from the cache. The suffix is always recomputed because it attends to every visual token, and its last token yields the embedding. The LM approximates in that unchanged visual tokens keep their cached keys and values even when earlier tokens have changed.

\textbf{Refresh.} Reused keys and values drift from those of a full encoding. We therefore re-encode all kept blocks, setting $\mathcal{C}_t \leftarrow \mathcal{K}_t$, when
\begin{equation}
|\mathcal{C}_t| > \rho\,|\mathcal{K}_t| \quad\text{or}\quad t - t_r \ge R,
\label{eq:refresh}
\end{equation}
where $t_r$ is the last refreshed screenshot, $\rho = 0.9$, and $R = 10$. The first condition catches a new screen, on which partial recomputation saves little, and the second bounds the drift.

\subsection{Consistency Adapter}
\label{sec:adapter}

We fine-tune the embedding model with LoRA~\citep{hu2022lora} on the training split of \bench (\S\ref{sec:bench-labels}) and use it to embed questions. Pruning and reuse, however, shift the embedding of a reduced screenshot away from that of the full one. We therefore train a consistency adapter, a second LoRA initialized from the first and used only to encode screenshots, by distilling the fine-tuned model $E_T$ on full screenshots into the adapter $E_S$ on pruned ones. For a screenshot $s$ and its pruned input $\tilde{s}$, the loss is
\begin{equation}
\mathcal{L} = 1 - \cos\bigl(E_S(\tilde{s}), E_T(s)\bigr) + \mathrm{KL}\bigl(p_T \,\|\, p_S\bigr) + \mathcal{L}_{\mathrm{margin}},
\label{eq:adapter}
\end{equation}
where $p_T$ and $p_S$ are softmax distributions of the teacher's and student's similarities over a question's candidate screenshots, and $\mathcal{L}_{\mathrm{margin}}$ is Margin-MSE loss on a relevant screenshot and a hard negative.

Following multi-objective distillation~\citep{sanh2019distilbert, jiao2020tinybert}, the three terms match the teacher's embedding~\citep{hwang2025d}, its ranking over candidates~\citep{abbasi2020compress, fang2021seed}, and its score gap on hard pairs~\citep{hofstatter2020improving}. Merged into the screenshot encoder's weights, the adapter adds no computation.
\section{The \bench Benchmark}
\label{sec:benchmark}
\vspace{-1mm}

Evaluating screenshot retrieval under real-time indexing (\S\ref{sec:background}) requires a screenshot history, questions about its past, and relevance labels. The closest benchmark~\citep{wang2025mirix} asks questions over a person's screenshot history, but scores generated answers without relevance labels, and its data is not public.

We therefore build \bench (\benchshort), with 10,426 test questions (plus 1,315 for training and validation) over three screenshot histories, two from human workers and one from a computer-use agent (Table~\ref{tab:benchmark}). Each question is labeled with the task recordings that answer it. \benchshort\ consists of three sets, \benchA, \benchB, and \benchC. The appendix gives further details.

\vspace{-1mm}
\subsection{Screen Histories}
\label{sec:bench-streams}

\textbf{Histories from human workers.} \benchA and \benchB each hold one worker's task recordings, in time order, from \citet{psai_computer_use_2025}, a public dataset of instructed tasks on websites and desktop applications on Windows. Screenshot are sampled every $\Delta = 5$\,s at $1280{\times}720$.

\textbf{History from a computer-use agent.} \benchC extends the \benchshort to a computer-use agent's screen. This reflects an emerging workflow: agents perform tasks, and humans later verify what they did; the history can also serve as the agent's own memory. We run GPT-6.1-sol~\citep{openai2026gpt61sol}, a frontier model, on the 184 tasks of myPCBench~\citep{jang2026mypcbench}. The tasks are built around one persona, whose Ubuntu virtual machine holds personal files, 17 self-hosted web applications, and desktop applications. Since each task involves many steps, we use an LLM to split them into 787 sub-tasks, which we then place on a 43-day calendar. Each sub-task forms one recording, captured at the same resolution and interval as the human histories.


\begin{table}[t]
\centering
\small
\setlength{\tabcolsep}{4pt}
\renewcommand{\arraystretch}{1.0}
\vspace{-2mm}
\caption{Statistics of the three screenshot histories of \benchshort.}
\vspace{1mm}
\label{tab:benchmark}
\begin{tabular*}{\columnwidth}{@{\extracolsep{\fill}}lccc@{}}
\toprule
 & \benchA & \benchB & \benchC \\
\midrule
Recorded from & human & human & agent \\
Days & 117 & 57 & 43 \\
Recordings & 2,105 & 946 & 787 \\
Screenshots & 63.5k & 12.8k & 55.9k \\
Questions & 5,111 & 3,507 & 1,808 \\
\quad single / multi & 4,265 / 846 & 3,096 / 411 & 1,654 / 154 \\
\bottomrule
\end{tabular*}
\vspace{-6mm}
\end{table}

\subsection{Questions}
\label{sec:bench-questions}

\benchshort\ has single- and multi-recording questions $q$, answered by one ($n = 1$) or several ($n \geq 2$) recordings (Table~\ref{tab:qtypes}).

\textbf{Single-recording questions.} A single-recording question asks where, when, what, or how a task was done, or which value was entered, and is phrased in the first person, as the worker or the persona would ask it. Opus 5.5~\citep{anthropic2026opus55} writes one of each type for each recording, based on its annotations, which describe the task, application, and time. The generator never sees screenshots or OCR text, so it favors neither index type. Most screenshots in every history show a clock, so times can be matched on screen.

To keep each answer unique, the same model, as a judge, compares each question with similar recordings, discards it if another recording answers differently, and labels recordings with the same answer as relevant. For \benchC, GPT-6-astra~\citep{openai2026gpt6astra} also checks that the recording's screens and actions support the answer.

\textbf{Multi-recording questions.} A multi-recording question counts how often an event occurred, compares two values, orders two events, or compares two durations. Its answer is computed from the annotations or verified answers of single recordings, and an LLM only phrases the question. To ensure each question needs all its evidence, a comparison, order, or duration, a question is kept only if an LLM answers it correctly from the relevant recordings, but fails once any one is removed. A count must match two independent LLM recounts exactly. In \benchC, GPT-6-astra also checks that every recording a question combines shows its task done on screen.

\begin{table}[t!]
\centering
\small
\renewcommand{\arraystretch}{1.3}
\vspace{-2mm}
\caption{Question types in \benchshort, with examples.}
\label{tab:qtypes}
\begin{tabularx}{\columnwidth}{@{}l>{\raggedright\arraybackslash}X@{}}
\toprule
Type & Example question and answer \\
\midrule
\multicolumn{2}{@{}l}{\textbf{Single-recording}} \\
\emph{where} & Where did I look up highly rated Vienna hotels with parking and breakfast? \ans{Booking.com} \\
\emph{when} & When did I check the airline's size rules for flying with a bike? \ans{Jul 29, 4:44\,pm} \\
\emph{what} & What did I do on Rent.com on the afternoon of August 7? \ans{Checked utility costs} \\
\emph{how} & How did I create the battery report? \ans{Ran \texttt{powercfg}} \\
\emph{value} & What return date did I set for the JFK round-trip flight? \ans{June 20th} \\
\midrule
\multicolumn{2}{@{}l}{\textbf{Multi-recording}} \\
\emph{count} & How many times in September did I search for places to stay on Airbnb? \ans{3} \\
\emph{comparison} & Did I search for more nights for the Sydney hotel or for the Jakarta hotel? \ans{Sydney} \\
\emph{order} & Did I look up a pivot table tutorial first, or check the price of Bitcoin? \ans{Pivot table} \\
\emph{duration} & Which took me longer: finding hiking trails nearby, or looking up the symbol for gold? \ans{Hiking trails} \\
\bottomrule
\end{tabularx}
\vspace{-8mm}
\end{table}

\vspace{-1mm}
\subsection{Relevance Labels and Splits}
\label{sec:bench-labels}

\textbf{Relevance labels.} Questions ask about tasks, not single screenshots, so relevance is labeled per task recording. For a single-recording question ($n = 1$), $\mathcal{S}^{+}_1(q)$ in Eq.~\eqref{eq:recall} holds the screenshots of its recording and of every recording the judge labeled relevant. A multi-recording question has one set per piece of evidence, holding the screenshots of the recording that supplies it and of any recording judged to give the same evidence. As in parallel multi-hop search~\citep{ko2026hybrid}, each piece of evidence can be retrieved independently and then combined. Comparison, order, and duration questions have $n = 2$, and for a count question $n$ equals its answer, from 2 to 8. Since a \benchC recording averages 71 screenshots, we also score \benchC on only the screenshots that show the evidence.

\begin{figure*}[t!]
\centering
\includegraphics[width=\textwidth]{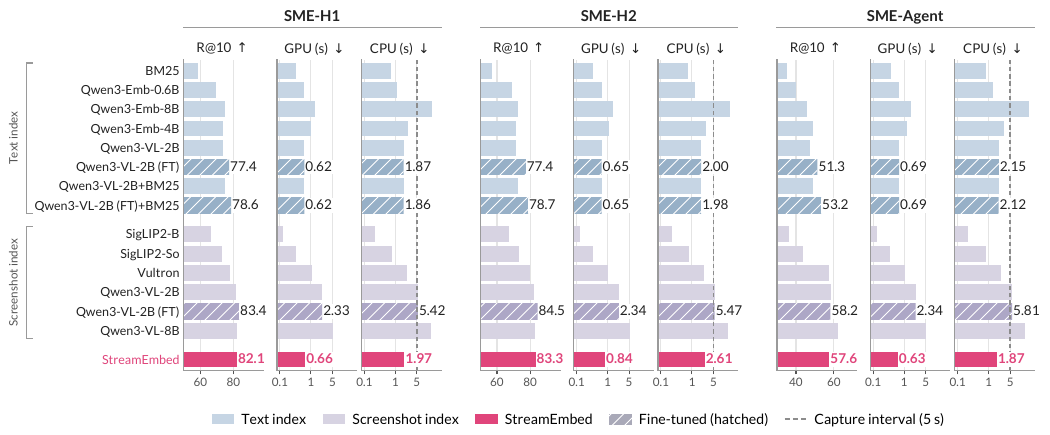}
\vspace{-7mm}
\caption{\textbf{Main results.} \method approaches the accuracy of a screenshot index at close to the latency of a text index. Recall@10 and mean indexing latency per screenshot on the GPU and CPU of an M2 Pro (\S\ref{sec:eval-setup}), latency on a log scale. Text indexes use Apple Vision OCR on the GPU and RapidOCR on the CPU, since Apple Vision OCR runs on the Neural Engine. OCR latency is almost equal to the BM25 latency, since BM25 adds under 0.1\,ms.}
\vspace{-3mm}
\label{fig:main}
\end{figure*}

\section{Experiments}
\label{sec:results}

We first compare \method with text and screenshot indexes (\S\ref{sec:eval-main}). We then index a long stream, simulating up to five years of screenshots to test how different index types scale (\S\ref{sec:eval-stream}). Finally, we apply \method to other encoders (\S\ref{sec:eval-encoders}) and ablate its components (\S\ref{sec:eval-ablation}).

\subsection{Setup}
\label{sec:eval-setup}

\textbf{Datasets.} We evaluate on the three histories of \benchshort\ (\S\ref{sec:benchmark}). Fine-tuned models are trained on 1,000 questions about the earliest recordings of \benchA, and all checkpoints and settings are selected on 315 questions about the next 79 recordings; neither set overlaps with the test questions. \benchA is thus in-domain for every fine-tuned model, \benchB tests transfer to another worker, and \benchC is out-of-domain, recorded from a computer-use agent on another operating system (\S\ref{sec:bench-streams}).

\textbf{Baselines.} We compare \method with text and screenshot indexes. A \textbf{text} index extracts the OCR text of each screenshot with Apple Vision OCR~\citep{applevision} for GPU latency and with RapidOCR~\citep{RapidOCR2021} for CPU latency, since Apple Vision OCR runs on the Neural Engine. It then ranks the text with BM25~\citep{robertson2009probabilistic}, or embeds it with Qwen3-Emb-0.6B, -4B, or -8B~\citep{zhang2025qwen3} (and gte-modernbert-base~\citep{li2023towards, zhang2024mgte} in Figure~\ref{fig:intro}) or with the embedding model Qwen3-VL-2B~\citep{li2026qwen3}. We also fuse the ranking of Qwen3-VL-2B with that of BM25 by reciprocal rank fusion~\citep{cormack2009reciprocal}. A \textbf{screenshot} index embeds the full screenshot with the NaFlex variants of SigLIP2-B and SigLIP2-So400m (SigLIP2-So)~\citep{tschannen2025siglip2}, VultronRetriever-Flash (Vultron)~\citep{vultron}, or Qwen3-VL-2B or -8B~\citep{li2026qwen3}. Qwen3-VL-2B serves as both a text and a screenshot encoder in the main comparison, used base or fine-tuned with LoRA (FT).

\textbf{Implementation.} \method uses $\tau = \delta = 8$, tiles of $3{\times}3$ blocks, $\rho = 0.9$, and $R = 10$, chosen on the \benchA validation questions, and encodes screenshots with the consistency adapter (\S\ref{sec:adapter}). 
The Qwen3-VL screenshot baselines run as released, with a Conv3d patch embedding. \method computes it with an equivalent linear one (\S\ref{sec:eval-ablation}).
LoRA and the consistency adapter are merged into the weights and do not change latency.

\textbf{Metrics.} We report (i) Recall@10 of each dataset for accuracy and (ii) indexing latency for cost, the mean time to encode a screenshot and add it to the index on the GPU and CPU of an Apple M2 Pro, including OCR for the text index. We measure latency over the first 50 consecutive screenshots, or all if fewer, of 30, 30, and 20 recordings sampled from \benchA, \benchB, and \benchC.

\begin{figure*}[t!]
  \centering
  \includegraphics[width=\textwidth]{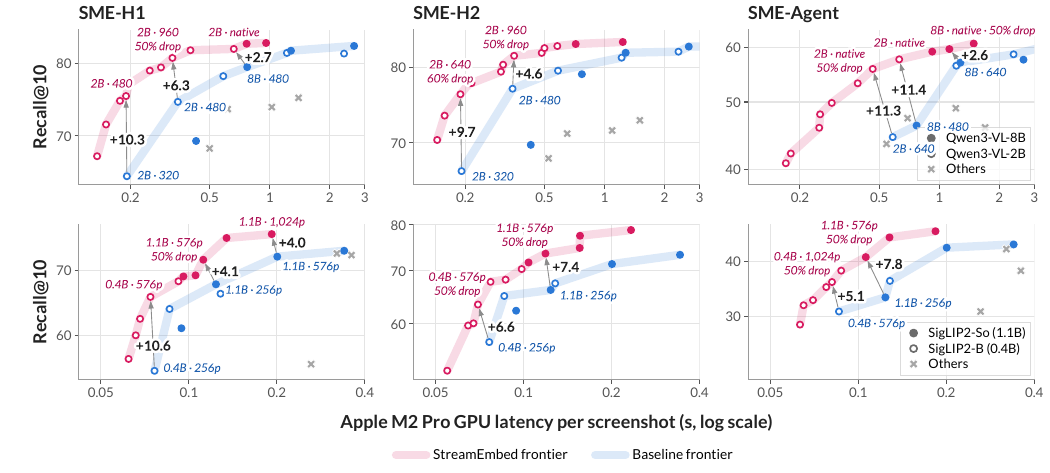}
  \vspace{-8mm}
  \caption{\textbf{\method with lower resolution and smaller encoders.} Bands are the Pareto frontiers of Qwen3-VL (top) and SigLIP2 (bottom) over encoder size and resolution, with and without \method; arrows give the R@10 gain of \method at a similar or lower latency. Labels give the encoder size and input width (SigLIP2: maximum patches, p) and, for \method, the fraction of dropped blocks. Gray crosses are other indexes, off both frontiers. 
  }
  \vspace{-3mm}
  \label{fig:frontier}
\end{figure*}

\subsection{Main Results}
\label{sec:eval-main}

Figure~\ref{fig:main} compares \method against all text and screenshot indexes across the three histories, reporting R@10 and mean latency on the GPU and CPU.

\textbf{\method is 2.8--3.7$\times$ faster than the screenshot index and stays within 1.3 R@10 of it.} Against the fine-tuned Qwen3-VL-2B screenshot index, which takes 2.33--2.34\,s per screenshot on the GPU, \method takes 0.66, 0.84, and 0.63\,s on \benchA, \benchB, and \benchC and scores 1.3, 1.2, and 0.6 R@10 lower (Figure~\ref{fig:main}). It is more accurate than the base screenshot index on \benchA and \benchB and 1.0 R@10 below it on \benchC. Only Qwen3-VL-8B is clearly more accurate, on \benchC, at about eight times the latency.

\textbf{\method is more accurate than every text index, at a similar latency.} The strongest text index, the fine-tuned Qwen3-VL-2B fused with BM25, scores 3.5--4.6 R@10 below \method, and larger text encoders such as Qwen3-Emb-8B do not close this gap while taking longer. On the GPU, \method nearly matches its latency on \benchA (0.66 vs.\ 0.62\,s) and is faster on \benchC (0.63 vs.\ 0.69\,s). On \benchB it takes 0.84 against 0.65\,s, but it still fits within a capture interval of 1\,s, which the screenshot index (2.34\,s) does not. The same holds on the CPU. \method takes 1.87--2.61\,s, close to the strongest text index (1.86--2.12\,s), while the fine-tuned screenshot index takes 5.42--5.81\,s, longer than even a 5\,s capture interval.

\textbf{\method moves the Pareto frontier beyond lower resolution and smaller encoders.} A common way to cut indexing latency is to lower the input resolution or to use a smaller encoder. Whereas Figure~\ref{fig:intro} runs each baseline at its default resolution, Figure~\ref{fig:frontier} revisits the comparison by applying both to the screenshot index as well as to \method, which can also raise its drop ratio. Even then, on both encoder families and all three histories, \method scores 2.6--11.4 R@10 higher than the baseline at a similar or lower latency. The gain is largest at low latency, where the baseline must fall back to a small input. At 0.19\,s on \benchA, Qwen3-VL-2B with \method at 480$\times$270 reaches 75.3 R@10, against 65.0 for the full screenshot at 320$\times$180. The gain narrows near 0.7\,s, where both sides approach the accuracy of the full encoder, yet \method on Qwen3-VL-2B still scores 2.7 R@10 above Qwen3-VL-8B at 480$\times$270 while running faster. Lowering the resolution or the encoder size applies to \method as well, and \method makes each resolution cheaper, so most points on its frontier pair \method with a lower resolution.


\begin{table*}[t]
  \centering
  \small
  \renewcommand{\arraystretch}{1.0}
  \setlength{\tabcolsep}{2pt}
  \vspace{-2mm}
  \caption{\textbf{Index types as the index grows.} B/entry is RAM per screenshot. R@10 is on real histories, and subscripts give the change from Flat. RAM, search, and insertion on CPU are for 1--5 years of screenshots (2.1M/year), simulated from the real embeddings with Gaussian noise. PQ and 1-bit re-rank the top 1,000 with fp16 vectors on the SSD. Search times the index only; encoding a query adds 52\,ms on average. -- exceeds 24\,GB of RAM.
  }
  \label{tab:index-scaling}
  \begin{tabular*}{\textwidth}{@{\extracolsep{\fill}}l c cc ccc ccc ccc@{}}
      \toprule
       & & \multicolumn{2}{c}{R@10} & \multicolumn{3}{c}{RAM (GB)} & \multicolumn{3}{c}{Search (ms)} & \multicolumn{3}{c}{Insert (ms)} \\
      \cmidrule(lr){3-4} \cmidrule(lr){5-7} \cmidrule(lr){8-10} \cmidrule(l){11-13}
      Index & B/entry & \benchA & \benchC & 1\,yr & 2\,yr & 5\,yr & 1\,yr & 2\,yr & 5\,yr & 1\,yr & 2\,yr & 5\,yr \\
      \midrule
      Flat & 4,096 & 82.1 & 57.6 & 8.6 & 17.2 & -- & 440 & 881 & -- & \textbf{0.002} & \textbf{0.002} & -- \\
      IVF & 4,104 & 80.9\rlap{\textsubscript{$-$1.2}} & 54.8\rlap{\textsubscript{$-$2.8}} & 8.6 & 17.3 & -- & 92 & 184 & -- & 0.09 & 0.09 & -- \\
      HNSW & 4,368 & 81.7\rlap{\textsubscript{$-$0.4}} & 56.2\rlap{\textsubscript{$-$1.4}} & 9.2 & 18.4 & -- & \textbf{0.96} & \textbf{1.3} & -- & 0.55 & 0.67 & -- \\
      PQ (128\,B) & 128 & 82.0\rlap{\textsubscript{$-$0.1}} & 57.5\rlap{\textsubscript{$-$0.1}} & 0.27 & 0.54 & 1.35 & 65 & 129 & 359 & 0.14 & 0.14 & 0.14 \\
      1-bit (all 2048 dims) & 256 & \textbf{82.1} & \textbf{57.6} & 0.54 & 1.08 & 2.69 & 24.8 & 44.1 & 102 & \textbf{0.007} & \textbf{0.007} & \textbf{0.007} \\
      1-bit (first 512 dims) & \textbf{64} & \textbf{82.1} & \textbf{57.6} & \textbf{0.13} & \textbf{0.27} & \textbf{0.67} & 15.5 & 29.5 & \textbf{56.0} & 0.010 & 0.010 & 0.010 \\
      \bottomrule
  \end{tabular*}
\vspace{-5.5mm}
\end{table*}

\begin{table}[t]
\centering
\scriptsize
\renewcommand{\arraystretch}{1.08}
\setlength{\tabcolsep}{1.5pt}
\caption{\textbf{\method on other encoders.} Full screenshot and \method for each encoder, all fine-tuned with LoRA on \benchA. Latency is GPU\,/\,CPU in seconds per screenshot. 
Darker blue marks higher R@10 and darker pink marks lower GPU latency, within each column.
}
\label{tab:encoders}
\begin{tabular*}{\columnwidth}{@{\extracolsep{\fill}}l cc cc cc@{}}
\toprule
 & \multicolumn{2}{c}{\benchA} & \multicolumn{2}{c}{\benchB} & \multicolumn{2}{c}{\benchC} \\
\cmidrule(lr){2-3}\cmidrule(lr){4-5}\cmidrule(l){6-7}
 & R@10 & Latency & R@10 & Latency & R@10 & Latency \\
\midrule
Qwen3-VL-8B & \cellcolor{accF}84.0 & \cellcolor{latA}5.15\,/\,14.82 & \cellcolor{accF}85.0 & \cellcolor{latA}5.19\,/\,14.88 & \cellcolor{accF}61.9 & \cellcolor{latA}5.16\,/\,14.68 \\
\quad \method & \cellcolor{accF}83.0 & \cellcolor{latB}1.59\,/\,6.31 & \cellcolor{accF}83.0 & \cellcolor{latB}2.08\,/\,7.78 & \cellcolor{accF}60.7 & \cellcolor{latB}1.49\,/\,5.93 \\
\midrule
Qwen3-VL-2B & \cellcolor{accF}83.4 & \cellcolor{latB}2.33\,/\,5.42 & \cellcolor{accF}84.5 & \cellcolor{latB}2.34\,/\,5.47 & \cellcolor{accF}58.2 & \cellcolor{latB}2.34\,/\,5.81 \\
\quad \method & \cellcolor{accF}82.1 & \cellcolor{latD}0.66\,/\,1.97 & \cellcolor{accF}83.3 & \cellcolor{latC}0.84\,/\,2.61 & \cellcolor{accE}57.6 & \cellcolor{latD}0.63\,/\,1.87 \\
\midrule
SigLIP2-So (1.1B) & \cellcolor{accD}77.5 & \cellcolor{latE}0.34\,/\,0.81 & \cellcolor{accD}78.7 & \cellcolor{latE}0.34\,/\,0.80 & \cellcolor{accC}49.6 & \cellcolor{latE}0.34\,/\,0.81 \\
\quad \method & \cellcolor{accD}76.3 & \cellcolor{latE}0.19\,/\,0.39 & \cellcolor{accD}78.8 & \cellcolor{latE}0.23\,/\,0.51 & \cellcolor{accC}46.9 & \cellcolor{latE}0.18\,/\,0.37 \\
\midrule
Vultron (0.8B) & \cellcolor{accE}78.9 & \cellcolor{latC}1.05\,/\,2.47 & \cellcolor{accE}80.7 & \cellcolor{latC}1.05\,/\,2.51 & \cellcolor{accF}60.4 & \cellcolor{latC}1.05\,/\,2.44 \\
\quad \method & \cellcolor{accD}77.8 & \cellcolor{latC}0.71\,/\,1.22 & \cellcolor{accD}79.7 & \cellcolor{latC}0.79\,/\,1.49 & \cellcolor{accE}56.8 & \cellcolor{latD}0.67\,/\,1.21 \\
\midrule
SigLIP2-B (0.4B) & \cellcolor{accA}68.7 & \cellcolor{latF}0.13\,/\,0.22 & \cellcolor{accA}71.4 & \cellcolor{latF}0.13\,/\,0.22 & \cellcolor{accA}39.4 & \cellcolor{latF}0.13\,/\,0.22 \\
\quad \method & \cellcolor{accA}68.2 & \cellcolor{latF}0.09\,/\,0.12 & \cellcolor{accA}70.3 & \cellcolor{latF}0.10\,/\,0.15 & \cellcolor{accA}38.1 & \cellcolor{latF}0.09\,/\,0.12 \\
\bottomrule
\end{tabular*}
\vspace{-3mm}
\end{table}

\subsection{Indexing a Long Stream}
\label{sec:eval-stream}

So far, we have shown that \method\ makes real-time encoding practical, but on-device retrieval poses a second challenge: can a growing screenshot index fit within a laptop's RAM? Memory already constrains deployment (\eg, Recall~\citep{microsoftrecall} requires at least 16\,GB).

To quantify how far an index can go, we grow a simulated index from one to five years of screenshots (2.1M per year at one every 5\,s over eight hours a day, whose fp16 embeddings take 8.6\,GB) by adding Gaussian noise to copies of the real embeddings. We compare exact search (Flat), approximate search with an inverted file (IVF)~\citep{jegou2011product} or a graph (HNSW)~\citep{malkov2020efficient}, and compressed codes kept in RAM whose top 1,000 candidates are re-ranked with the full embeddings on disk, namely product quantization (PQ)~\citep{jegou2011product} and 1-bit codes~\citep{shakir2024quantization, yamada2021efficient}, all implemented in Faiss~\citep{douze2024faiss}.

Table~\ref{tab:index-scaling} reports the accuracy of each index on \benchA and \benchC, and its RAM, search time, and insertion time as the history grows, on the CPU of the M2 Pro.

\textbf{Indexes that keep full embeddings in RAM may not scale to years of screenshots.} Flat needs 8.6\,GB of RAM per year and exceeds 24\,GB at five years, and its search time grows with the index, from 440\,ms at one year to 881\,ms at two. IVF searches about five times faster than Flat but loses up to 2.8 R@10, since relevant screenshots can fall outside the clusters it searches. HNSW searches in about 1\,ms but needs 7\% more RAM than Flat for its graph and loses up to 1.4 R@10.

\textbf{Compressed codes with re-ranking from disk keep R@10 at a small fraction of the RAM.} PQ and 1-bit codes stay within 0.1 R@10 of Flat, since the codes only need to place the relevant screenshots among the top 1,000, and re-ranking with the full embeddings restores their order. 1-bit codes of the first 512 dimensions take 64 bytes per screenshot, 64 times less than Flat, and hold five years in 0.67\,GB. Comparing bit codes is also cheap, so they search in 56\,ms after five years, against 359\,ms for PQ. With the 52\,ms of encoding a question, retrieval takes about 0.1\,s even after five years.

\textbf{Insertion stays negligible even with five years of history.} Inserting a screenshot takes at most 0.67\,ms, for HNSW, and 0.01\,ms for 1-bit codes, against 0.63--0.84\,s to encode it with \method. This confirms the premise of \S\ref{sec:challenges} that the cost of real-time indexing lies in encoding, not insertion, even as the history grows, and it is this cost that \method reduces.

\begin{table}[t]
\centering
\scriptsize
\renewcommand{\arraystretch}{1.08}
\setlength{\tabcolsep}{1.5pt}
\caption{\textbf{Ablation.} Components of \method added one at a time to the fine-tuned Qwen3-VL-2B screenshot index. Latency is GPU\,/\,CPU in sec per screenshot. 
}
\label{tab:ablation}
\begin{tabular*}{\columnwidth}{@{\extracolsep{\fill}}l cc cc cc@{}}
\toprule
 & \multicolumn{2}{c}{\benchA} & \multicolumn{2}{c}{\benchB} & \multicolumn{2}{c}{\benchC} \\
\cmidrule(lr){2-3}\cmidrule(lr){4-5}\cmidrule(l){6-7}
 & R@10 & Latency & R@10 & Latency & R@10 & Latency \\
\midrule
Screenshot (FT) & \cellcolor{accF}83.4 & \cellcolor{latA}2.33\,/\,5.42 & \cellcolor{accF}84.5 & \cellcolor{latA}2.34\,/\,5.47 & \cellcolor{accF}58.2 & \cellcolor{latA}2.34\,/\,5.81 \\
+ Linear patch embed. & \cellcolor{accF}83.4 & \cellcolor{latB}1.65\,/\,5.42 & \cellcolor{accF}84.5 & \cellcolor{latC}1.64\,/\,5.47 & \cellcolor{accF}58.2 & \cellcolor{latB}1.66\,/\,5.81 \\
+ Spatial pruning & \cellcolor{accD}82.8 & \cellcolor{latE}1.00\,/\,2.96 & \cellcolor{accD}83.8 & \cellcolor{latE}1.17\,/\,3.64 & \cellcolor{accD}57.1 & \cellcolor{latD}1.08\,/\,3.48 \\
\quad w/o grid keep & \cellcolor{accC}82.6 & \cellcolor{latE}0.84\,/\,2.49 & \cellcolor{accC}83.5 & \cellcolor{latE}1.03\,/\,3.07 & \cellcolor{accA}55.2 & \cellcolor{latE}0.91\,/\,2.60 \\
+ Temporal reuse & \cellcolor{accA}81.8 & \cellcolor{latF}0.66\,/\,1.97 & \cellcolor{accA}82.9 & \cellcolor{latF}0.84\,/\,2.61 & \cellcolor{accC}56.2 & \cellcolor{latF}0.63\,/\,1.87 \\
+ Consistency adapter & \cellcolor{accB}82.1 & \cellcolor{latF}0.66\,/\,1.97 & \cellcolor{accB}83.3 & \cellcolor{latF}0.84\,/\,2.61 & \cellcolor{accE}57.6 & \cellcolor{latF}0.63\,/\,1.87 \\
\bottomrule
\end{tabular*}
\vspace{-5mm}
\end{table}

\vspace{-1mm}
\subsection{Generalization to Other Encoders}
\label{sec:eval-encoders}

Table~\ref{tab:encoders} applies \method with the same thresholds to four other screenshot encoders, each fine-tuned with LoRA on \benchA, and compares it with each encoder on full screenshots.

\textbf{\method generalizes to every encoder, staying close to its full-screenshot accuracy.} It lowers the GPU latency of all five encoders, by 1.3--1.9$\times$ on SigLIP2 and Vultron and by 2.5--3.7$\times$ on Qwen3-VL, while losing only 1.3 R@10 on average.

\textbf{The speedup grows with the computation that reuse can skip.} \method speeds up the Qwen3-VL encoders by 2.5--3.7$\times$ on the GPU, since temporal reuse also saves computation in the language model. On SigLIP2, which has no language model, and on Vultron, whose language model mostly uses linear attention, \method reuses keys and values only in the vision encoder and still speeds them up by 1.3--1.9$\times$ on the GPU and 1.5--2.2$\times$ on the CPU. On Vultron, which stores one vector per visual token, pruning also stores 23--30\% fewer vectors per screenshot, trading some accuracy on \benchC for a smaller index.

\subsection{Ablation}
\label{sec:eval-ablation}

Table~\ref{tab:ablation} adds the components of \method one at a time to the released Qwen3-VL-2B, fine-tuned with LoRA, starting with the equivalent linear patch embedding that \method uses for Qwen3-VL (\S\ref{sec:eval-setup}).

\textbf{\method speeds up encoding by 2.8--3.7$\times$ over the released model, mostly through spatial pruning and temporal reuse.} Spatial pruning removes the repetition within a screenshot and temporal reuse removes the repetition across screenshots, each speeding up encoding by about 1.5$\times$ at about 1 R@10. Their savings thus multiply, to 2.0--2.6$\times$ together. Their overhead is small, as pruning and change detection add about 0.01\,s per screenshot, and in the final configuration the vision encoder and the language model each run two to three times faster than without them.

\textbf{Grid keep and the consistency adapter recover the accuracy lost due to pruning on \benchC.} Pruning every redundant block costs far more on \benchC than on the human histories, suggesting that dropping every redundant block shifts the embedding. Grid keep recovers most of this loss for about 0.15\,s per screenshot, and the consistency adapter recovers more at no latency cost, leaving \method within 1.3 R@10 of the full screenshot on all three histories.

\vspace{-1mm}
\section{Related Work}
\label{sec:related_work}

\vspace{-1mm}
\paragraph{Screen history.}
Re-finding what was once on a computer screen has long been studied, with people searching the files, e-mail, and web pages they have opened~\citep{dumais2003stuff} or browsing past screenshots~\citep{hu2020screentrack}. 
Such histories serve not only people but also agents, which complete more later tasks when they reuse past experience~\citep{wang2024agent, agashe2025agent} or see past screenshots rather than summaries~\citep{luo2026causalcache}. 
Recent systems build such histories using vision-language models (VLMs), turning screenshots into the text memory of an agent~\citep{wang2025mirix} or a log of each task~\citep{yin2026focal}, or answering questions over them~\citep{li2025omniquery}. 
These systems rely on large cloud models~\citep{wang2025mirix, li2025omniquery} or, to run on the device, process only some screenshots~\citep{yin2026focal}, and either way represent screenshots as text. 
Yet retrieving from pixels is more accurate than from text on documents and the web~\citep{ma2024unifying, faysse2025colpali}.
We thus keep every screenshot as an embedding of its pixels, computed on the device.

\vspace{-2mm}
\paragraph{Growing indexes.}
Earlier work has studied indexes that grow over time, but it rarely treats the cost of encoding each new item as the bottleneck. Real-time search indexes text, which is cheap \citep{busch2012earlybird}, and updatable vector indexes leave embedding as a separate step \citep{xu2023spfresh}. Work on evolving corpora studies how to add documents to a trained retriever~\citep{kishore2023incdsi} and when to update it~\citep{ko2025should, kim2024dynamicer}. For screenshots, however, encoding is the bottleneck. Work that faces this cost encodes less or defers it, using cheaper models~\citep{hsieh2018focus}, fewer frames~\citep{kang2022tasti}, early exit~\citep{cai2025reminisce}, smaller encoders~\citep{teiletche2026modernvbert}, or lower resolution~\citep{wang2026pixelrag}. Instead, we remove only the computation that screens repeat, which is combined with a smaller encoder, lower resolution, or a skip gate. A growing index must also stay small. \citet{lyu2026frustratingly} re-rank compressed codes from disk, and \citet{wang2026leann} recompute embeddings at search time on a laptop.
We study how to fit a screenshot index in a laptop's memory.

\vspace{-2mm}

\paragraph{Streaming VLMs.}
Streaming VLMs keep each arriving frame in their context to answer questions about the video at any time~\citep{chen2024videollm}, so their main concern is what to keep.
Instead, we store each screenshot in an index for later queries, as in~\citet{jeong2025videorag}, so our main concern is how fast each one is encoded. 
This leads to three differences.
These models keep tokens or their keys and values, which they evict, compress~\citep{xu2026streamingvlm}, or retrieve for each question~\citep{di2025streaming, ning2026livevlm}.
Our index holds only one embedding per screenshot.
Methods that cut computation across frames decide what to skip inside or after the vision encoder~\citep{yao2025timechat, wang2026accelerating}.
Others decide from pixels but reuse computation in at most one of the encoder and the LM~\citep{choudhury2024dont, xu2026vla}.
In contrast, our method decides fast from pixels before the encoder and reuses unchanged blocks in both.
Finally, prior work keeps accuracy with modules added at inference~\citep{aggarwal2026vtr, hwang2025d, soldan2025residualvit}.
Our adapter is instead a LoRA merged into the weights, adding no latency.

\section{Conclusion}
\label{sec:conclusion}

We introduced \method, which indexes a growing screenshot history in real time by encoding it as a stream. Since a rendered screen repeats most of its blocks, \method prunes blocks that repeat a neighbor, reuses keys and values of unchanged blocks in both the vision encoder and the language model, and keeps the reduced embedding close to the full one at no added cost. On \bench, built from the screenshot histories of human workers and a computer-use agent, \method matches a text index in latency and a screenshot index in accuracy, while binary codes keep a year of screenshots within the laptop's memory. Future work could learn what to recompute from the stream itself, lowering cost and raising accuracy further. Agents could also search a person's screenshot history to complete long-horizon tasks that depend on screens seen days or weeks earlier.

\section*{Acknowledgments}

We sincerely thank Donghyun Lee, Prasann Singhal, and Ryan Wang for their valuable feedback and support.

\bibliography{example_paper}
\bibliographystyle{mlsys2025}

\clearpage
\appendix
\section{Benchmark Details}
\label{app:benchmark}

Table~\ref{tab:bench-compare} compares \bench with ScreenshotVQA~\citep{wang2025mirix}, a screen benchmark.


\textbf{Questions.} The generator writes questions for 12 recordings at a time from their annotations, which are the instruction, application, times, and planned steps for \benchA and \benchB, and the sub-task instruction, the agent's actions, and the scheduled time for \benchC. To keep answers unique, the judge compares each question with up to 30 recordings of the same history chosen from annotations alone, such as recordings of the same task or on the same site, and always every sibling sub-task in \benchC. No evaluated index chooses them, so the labels favor neither index type. Each count recount (\S\ref{sec:bench-questions}) runs over a shuffled list of the candidate recordings.

\textbf{Relevance and splits.} For screenshot-level scoring on \benchC, GPT-6-astra marks the keyframes that show each question's evidence, and each keyframe counts with its run of near-identical screenshots. Training uses the earliest 710 recordings of \benchA, and the 1,121 questions about the 433 of them without training questions are test questions. Training and validation recordings stay in the index and are relevant to 54 test questions, and removing these questions changes R@10 by at most 0.03 for both the fine-tuned full screenshot and \method.

\textbf{Data.} \benchA and \benchB come from \citet{psai_computer_use_2025}, recorded by consenting workers who authorized public release, under the MIT license. \benchC shows a synthetic persona and holds no data of real people. We release all three histories with their questions, relevance labels, and splits.

\begin{table}[t!]
\centering
\small
\renewcommand{\arraystretch}{1.1}
\setlength{\tabcolsep}{1.5pt}
\caption{\textbf{\bench and ScreenshotVQA.} Days counts days with at least one recording, as in Table~\ref{tab:benchmark}, and Span counts calendar days from the first to the last recording, on the simulated calendar for \benchC. Screen time is the number of screenshots times the 5\,s interval. Relevance marks recording-level relevance labels, and Role is the use of each history when encoders are fine-tuned. For ScreenshotVQA, span and screenshots are averaged over its three histories.}
\label{tab:bench-compare}
\resizebox{\columnwidth}{!}{%
\begin{tabular}{@{}lcccc@{}}
\toprule
 & & \multicolumn{3}{c}{\bench (Ours)} \\
\cmidrule(l){3-5}
 & \begin{tabular}{@{}c@{}}Screenshot-\\VQA\end{tabular} & \benchA & \benchB & \benchC \\
\midrule
\multicolumn{5}{@{}l}{\textbf{History}} \\
Days & -- & 117 & 57 & 43 \\
Span (days) & 22 & 172 & 78 & 43 \\
Screenshots & 9.8k & 63.5k & 12.8k & 55.9k \\
\quad per recording & -- & 30 & 13 & 71 \\
Screen time (h) & -- & 88 & 18 & 78 \\
\midrule
\multicolumn{5}{@{}l}{\textbf{Questions}} \\
Questions & 87 & 5,111 & 3,507 & 1,808 \\
\midrule
\multicolumn{5}{@{}l}{\textbf{Evaluation}} \\
Relevance & \textcolor{gray}{\ding{55}} & \ding{51} & \ding{51} & \ding{51} \\
Scored on & answer & \multicolumn{3}{c}{retrieval} \\
Latency & -- & \multicolumn{3}{c}{indexing} \\
Role & -- & \begin{tabular}{@{}c@{}}train\\+ test\end{tabular} & \begin{tabular}{@{}c@{}}test\\(new worker)\end{tabular} & \begin{tabular}{@{}c@{}}test\\(new source)\end{tabular} \\
Released & \textcolor{gray}{\ding{55}} & \ding{51} & \ding{51} & \ding{51} \\
\bottomrule
\end{tabular}}
\end{table}

\section{Full Results}\label{app:results}

Table~\ref{tab:results-full} lists every system of the main results with R@10 split into single- and multi-recording questions, together with nDCG@10~\citep{jarvelin2002cumulated}. For a multi-recording question, R@10 is the fraction of its evidence sets hit, as in Eq.~\eqref{eq:recall}. nDCG@10 gives each evidence set one unit of gain at the rank of its first retrieved screenshot and normalizes by the ideal ranking of $\min(n, 10)$ sets.

\begin{table*}[t]
\centering
\footnotesize
\renewcommand{\arraystretch}{1.08}
\setlength{\tabcolsep}{2pt}
\caption{\textbf{Full main results.} R@10 over all questions (All), single-recording questions (Single), and multi-recording questions (Multi), and nDCG@10 over all questions. FT marks fine-tuning with LoRA on \benchA, and other rows use base models.}
\label{tab:results-full}
\begin{tabular*}{\textwidth}{@{\extracolsep{\fill}}l cccc cccc cccc@{}}
\toprule
 & \multicolumn{4}{c}{\benchA} & \multicolumn{4}{c}{\benchB} & \multicolumn{4}{c}{\benchC} \\
\cmidrule(lr){2-5}\cmidrule(lr){6-9}\cmidrule(l){10-13}
 & All & Single & Multi & nDCG & All & Single & Multi & nDCG & All & Single & Multi & nDCG \\
\midrule
\multicolumn{13}{@{}l}{\emph{Text index (Apple Vision OCR)}} \\
\quad BM25 & 58.6 & 64.3 & 29.9 & 51.0 & 56.7 & 58.6 & 42.3 & 47.1 & 35.3 & 36.0 & 27.3 & 26.2 \\
\quad Qwen3-Emb-0.6B & 69.2 & 75.0 & 39.9 & 61.2 & 69.3 & 72.1 & 48.8 & 60.1 & 40.1 & 41.1 & 29.6 & 28.3 \\
\quad Qwen3-Emb-4B & 73.8 & 80.0 & 42.3 & 66.2 & 71.5 & 75.0 & 45.3 & 63.0 & 49.0 & 50.4 & 34.1 & 36.0 \\
\quad Qwen3-Emb-8B & 75.0 & 81.1 & 44.5 & 68.2 & 72.8 & 76.0 & 49.1 & 64.4 & 46.0 & 47.3 & 31.1 & 33.9 \\
\quad Qwen3-VL-2B & 73.5 & 79.4 & 44.1 & 65.3 & 71.2 & 73.6 & 52.7 & 62.6 & 47.4 & 48.7 & 32.9 & 34.3 \\
\quad Qwen3-VL-2B (FT) & 77.4 & 83.3 & 47.7 & 70.4 & 77.4 & 80.1 & 57.2 & 68.4 & 51.3 & 52.9 & 34.1 & 37.4 \\
\quad Qwen3-VL-2B + BM25 & 74.8 & 80.5 & 45.9 & 66.5 & 72.4 & 74.7 & 54.8 & 63.5 & 48.8 & 50.1 & 34.8 & 35.2 \\
\quad Qwen3-VL-2B (FT) + BM25 & 78.6 & 84.4 & 49.4 & 71.3 & 78.7 & 81.1 & 60.0 & 68.7 & 53.2 & 54.7 & 37.0 & 38.3 \\
\midrule
\multicolumn{13}{@{}l}{\emph{Screenshot index}} \\
\quad SigLIP2-B & 66.2 & 73.0 & 31.9 & 58.0 & 67.5 & 70.6 & 44.1 & 57.2 & 36.1 & 37.1 & 25.3 & 26.2 \\
\quad SigLIP2-So & 73.3 & 80.2 & 38.8 & 65.6 & 73.3 & 76.3 & 51.2 & 63.4 & 43.8 & 45.4 & 26.1 & 31.7 \\
\quad Vultron & 77.8 & 84.7 & 43.4 & 70.3 & 79.9 & 83.4 & 53.4 & 69.4 & 57.6 & 60.0 & 31.5 & 43.1 \\
\quad Qwen3-VL-2B & 81.4 & 87.6 & 50.3 & 73.8 & 82.3 & 85.2 & 60.5 & 72.2 & 58.6 & 60.5 & 38.3 & 44.5 \\
\quad Qwen3-VL-2B (FT) & 83.4 & 89.2 & 54.2 & 76.3 & 84.5 & 87.3 & 63.7 & 74.7 & 58.2 & 59.9 & 40.9 & 44.9 \\
\quad Qwen3-VL-8B & 82.2 & 88.4 & 51.2 & 75.9 & 83.1 & 86.2 & 59.6 & 74.0 & 61.9 & 64.4 & 34.5 & 47.0 \\
\midrule
\method (Qwen3-VL-2B) & 82.1 & 88.0 & 52.2 & 75.5 & 83.3 & 86.1 & 62.4 & 73.9 & 57.6 & 59.4 & 39.0 & 43.8 \\
\bottomrule
\end{tabular*}
\end{table*}

\textbf{OCR engines.} Table~\ref{tab:ocr-engines} compares the text indexes on the two OCR engines, Apple Vision OCR used for GPU latency and RapidOCR used for CPU latency. The engine changes R@10 by at most 1.6, so every text index stays below \method with either engine.

\begin{table}[t]
\centering
\scriptsize
\renewcommand{\arraystretch}{1.08}
\setlength{\tabcolsep}{0pt}
\caption{\textbf{Text indexes on two OCR engines.} R@10 with OCR text from Apple Vision OCR or RapidOCR. FT marks fine-tuning with LoRA on \benchA.}
\label{tab:ocr-engines}
\begin{tabular*}{\columnwidth}{@{\extracolsep{\fill}}l ccc ccc@{}}
\toprule
 & \multicolumn{3}{c}{Apple Vision OCR} & \multicolumn{3}{c}{RapidOCR} \\
\cmidrule(lr){2-4}\cmidrule(l){5-7}
Text encoder & \benchA & \benchB & \benchC & \benchA & \benchB & \benchC \\
\midrule
Qwen3-Emb-0.6B & 69.2 & 69.3 & 40.1 & 70.0 & 69.7 & 39.8 \\
Qwen3-Emb-4B & 73.8 & 71.5 & 49.0 & 74.3 & 72.3 & 48.8 \\
Qwen3-Emb-8B & 75.0 & 72.8 & 46.0 & 74.6 & 71.2 & 46.3 \\
Qwen3-VL-2B & 73.5 & 71.2 & 47.4 & 73.5 & 71.7 & 46.8 \\
Qwen3-VL-2B (FT) & 77.4 & 77.4 & 51.3 & 78.0 & 77.3 & 51.7 \\
\bottomrule
\end{tabular*}
\end{table}

\section{Implementation Details}
\label{app:impl}

\textbf{Training.} Every fine-tuned encoder uses LoRA~\citep{hu2022lora} with rank 32 on the attention projections of the vision encoder and of the LM or text tower. It is trained with InfoNCE on the 1,000 training questions of \benchA, each with two hard negatives, for 300 steps at a learning rate of $7{\times}10^{-6}$ for Qwen3-VL, $5{\times}10^{-5}$ for SigLIP2, and $3{\times}10^{-5}$ for Vultron. The consistency adapter is a second LoRA initialized from the first and distilled from the frozen fine-tuned model with the loss of Eq.~\eqref{eq:adapter}, on screenshots pruned by the neighbor rule, for 300 steps at $10^{-5}$. Text indexes use Apple Vision OCR at its accurate level with language correction, and the fused text index combines Qwen3-VL-2B and BM25 by reciprocal rank fusion with $k = 10$ and weight 0.5 on BM25. All checkpoints, the settings of \method, and the fusion are chosen on the \benchA validation questions.

\textbf{Latency.} We time every index on an Amazon EC2 mac2-m2pro instance, an Apple M2 Pro with 32\,GB of memory, left idle, with PyTorch on MPS in fp16 for the GPU and in fp32 on 8 threads for the CPU. Screenshots are processed one at a time in their recorded order, caches start empty at each recording, and the timer runs from reading the image to the embedding, including OCR for a text index.

\textbf{Patch embedding.} The Qwen3-VL screenshot baselines run the released Conv3d patch embedding, and \method computes the same map as one matrix multiplication, whose embeddings have a cosine similarity of at least 0.99986 in fp16. Vultron runs with an equivalent but faster layout of its own Conv3d, so this baseline is faster than released.

\section{Resolution and Drop Variants}
\label{app:variants}

\begin{figure*}[t]
\centering
\includegraphics[width=\textwidth]{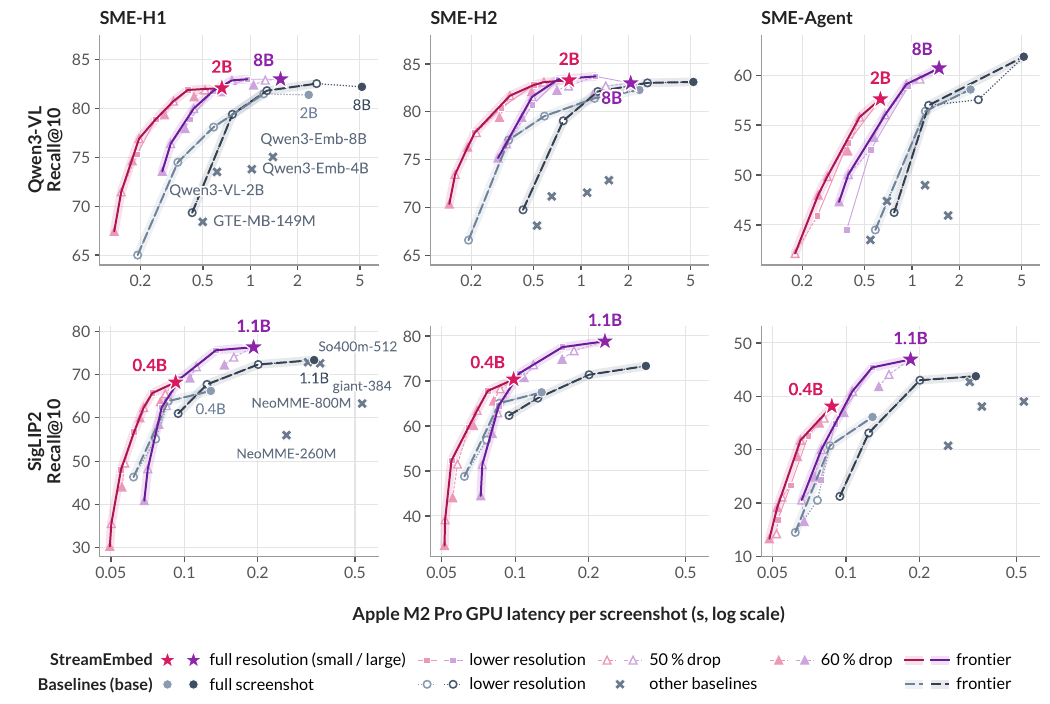}
\vspace{-5mm}
\caption{\textbf{Frontiers per model size.} R@10 against GPU latency per screenshot for Qwen3-VL (top) and SigLIP2 (bottom). Pink and purple mark \method on the small and large model, and light and dark slate the baselines. Solid lines are frontiers of \method over full resolution, lower resolutions, and drop rates of 50\% and 60\%, and dashed lines are frontiers of the base full screenshot over lower resolutions. Stars mark \method at full resolution. GTE-MB-149M is gte-modernbert-base on Apple Vision OCR text without language correction. In the Qwen3-VL row of \benchC, points below 42 R@10 are omitted.}
\label{fig:variants}
\end{figure*}

Figure~\ref{fig:variants} splits the frontiers of Figure~\ref{fig:frontier} by model size, so each size of \method is compared with the same size of the base encoder at lower resolutions. \method uses the fine-tuned encoder with its consistency adapter. Latency of the lower-resolution baselines is measured on \benchA and reused for the other histories.

\textbf{Fine-tuned baseline.} The baselines in Figures~\ref{fig:frontier} and~\ref{fig:variants} are base encoders. On Qwen3-VL, fine-tuning alone moves the full screenshot by $-0.4$ to 2.2 R@10 (Tables~\ref{tab:results-full} and~\ref{tab:encoders}), less than every Qwen3-VL gain in Figure~\ref{fig:frontier}. At $960{\times}540$, the fine-tuned Qwen3-VL-2B needs half the computation of a full screenshot and takes 1.22\,s per screenshot on the GPU with the released Conv3d, timed on \benchA, where full-screenshot latency matches the other histories. \method is 1.4--1.9$\times$ faster than it and 2.0 R@10 more accurate on \benchC, while staying within 1.3 R@10 on the human histories.

\section{Latency Breakdown}
\label{app:latency}

\begin{table}[t]
\centering
\small
\setlength{\tabcolsep}{3pt}
\renewcommand{\arraystretch}{1.08}
\caption{\textbf{Latency breakdown on \benchA.} Mean GPU latency per screenshot (s) of the fine-tuned Qwen3-VL-2B as parts of \method are added, with preprocessing in Total. The top block is the latency sample used throughout the paper, as in Table~\ref{tab:ablation}, with grid keep. The bottom block is a separate set of 2,000 \benchA test screenshots without grid keep. Comp.\ is computation in TFLOPs per screenshot.}
\label{tab:latency-breakdown}
\begin{tabular*}{\columnwidth}{@{\extracolsep{\fill}}l ccc cc@{}}
\toprule
 & \multicolumn{3}{c}{Latency (s)} & \multicolumn{2}{c}{Comp.} \\
\cmidrule(lr){2-4}\cmidrule(l){5-6}
 & Total & ViT & LM & ViT & LM \\
\midrule
\multicolumn{6}{@{}l}{\emph{Latency sample of the paper (994 screenshots)}} \\
Screenshot, Conv3d & 2.33 & 1.66 & 0.66 & 3.53 & 2.64 \\
+ Linear patch embed. & 1.65 & 0.96 & 0.68 & 3.53 & 2.64 \\
+ Spatial pruning & 1.00 & 0.52 & 0.45 & 1.87 & 1.60 \\
+ Temporal reuse & 0.66 & 0.34 & 0.29 & 1.01 & 0.85 \\
\midrule
\multicolumn{6}{@{}l}{\emph{Separate set of 2,000 screenshots, no grid keep}} \\
Screenshot, Conv3d & 2.35 & 1.67 & 0.66 & 3.53 & 2.64 \\
+ Linear patch embed. & 1.65 & 0.96 & 0.67 & 3.53 & 2.64 \\
+ Neighbor rule & 0.88 & 0.45 & 0.40 & 1.56 & 1.37 \\
+ Reuse in the ViT & 0.75 & 0.34 & 0.38 & 0.87 & 1.16 \\
+ Reuse in the LM & 0.67 & 0.34 & 0.30 & 0.87 & 0.75 \\
\bottomrule
\end{tabular*}
\end{table}

Table~\ref{tab:latency-breakdown} splits the GPU latency of \method into its vision encoder (ViT) and language model (LM). With reuse in the ViT only, the LM reuses the prefix before the first changed token. Pruning and reuse make the ViT 2.8$\times$ and the LM 2.3$\times$ faster than with the linear patch embedding alone, and the ViT 4.9$\times$ faster than in the released model. On the CPU of \benchA, the ViT runs 3.2$\times$ faster (3.72 to 1.16\,s) and the LM 2.2$\times$ faster (1.69 to 0.78\,s).

\textbf{Speedup and redundancy.} \method computes 1.95, 2.67, and 1.79 TFLOPs per screenshot on \benchA, \benchB, and \benchC, or 32\%, 43\%, and 29\% of the 6.17 TFLOPs of a full screenshot. This follows the redundancy of each history, where 72.4\%, 63.3\%, and 75.6\% of blocks repeat a neighbor or are unchanged (Appendix~\ref{app:redundancy}), and grid keep, refresh, and the suffix add a little work to the remaining 24--37\%. On the GPU, \method is accordingly 3.5$\times$, 2.8$\times$, and 3.7$\times$ faster than the released model. Its 99th-percentile latency, 1.47--1.77\,s, stays below the mean of the released full screenshot, 2.33--2.34\,s.

\textbf{Memory.} By our count from the model configuration, the cache of temporal reuse holds at most about 0.46\,GB in fp16 for Qwen3-VL-2B and does not grow with the history.

\section{Sensitivity to Thresholds}
\label{app:sensitivity}

Tables~\ref{tab:sens-grid} and~\ref{tab:sens-tol} show how R@10 and computation change around the settings of \method, varying one at a time. \method is not sensitive to its thresholds. Changing the refresh settings moves R@10 by at most 0.2 and the tile size by at most 1.4, and even raising both pixel tolerances 16-fold costs at most 2.0 R@10 while cutting computation up to 2.0$\times$. The settings are chosen on the \benchA validation questions (\S\ref{sec:eval-setup}), and the tables report R@10 on the test questions. Every row uses the fine-tuned Qwen3-VL-2B with all parts of \method, and computation is in TFLOPs per screenshot (TF).

\begin{table}[t]
\centering
\small
\setlength{\tabcolsep}{2.5pt}
\caption{\textbf{Grid keep tile size.} R@10 and computation (TF). None uses the neighbor rule alone, and -- was not run.}
\label{tab:sens-grid}
\begin{tabular*}{\columnwidth}{@{\extracolsep{\fill}}l cc cc cc@{}}
\toprule
 & \multicolumn{2}{c}{\benchA} & \multicolumn{2}{c}{\benchB} & \multicolumn{2}{c}{\benchC} \\
\cmidrule(lr){2-3}\cmidrule(lr){4-5}\cmidrule(l){6-7}
Tile & R@10 & TF & R@10 & TF & R@10 & TF \\
\midrule
None & 82.4 & 1.68 & 82.9 & 2.37 & 55.5 & 1.59 \\
$2{\times}2$ & 82.0 & 2.13 & 83.1 & 2.88 & 56.9 & 1.92 \\
$3{\times}3$ (\method) & 82.1 & 1.95 & 83.3 & 2.67 & 57.6 & 1.79 \\
$4{\times}4$ & -- & -- & -- & -- & 56.2 & 1.74 \\
\bottomrule
\end{tabular*}
\end{table}

\begin{table}[t]
\centering
\small
\setlength{\tabcolsep}{3pt}
\caption{\textbf{Pixel tolerances and refresh.} R@10 and computation (TF). The top block raises $\tau$ and $\delta$ together, and the bottom block changes one reuse setting on \benchB. -- was not run.}
\label{tab:sens-tol}
\begin{tabular*}{\columnwidth}{@{\extracolsep{\fill}}l cc cc@{}}
\toprule
 & \multicolumn{2}{c}{\benchB} & \multicolumn{2}{c}{\benchC} \\
\cmidrule(lr){2-3}\cmidrule(l){4-5}
 & R@10 & TF & R@10 & TF \\
\midrule
$\tau = \delta = 8$ (\method) & 83.6 & 2.67 & 57.4 & 1.79 \\
$\tau = \delta = 16$ & 83.0 & 2.34 & 56.4 & 1.59 \\
$\tau = \delta = 32$ & 83.1 & 1.99 & 55.8 & 1.31 \\
$\tau = \delta = 64$ & 82.8 & 1.82 & 55.8 & 1.04 \\
$\tau = \delta = 128$ & 82.6 & 1.55 & 55.4 & 0.88 \\
\midrule
$\delta = 16$ & 83.4 & 2.45 & -- & -- \\
$\delta = 32$ & 83.2 & 2.21 & -- & -- \\
$\delta = 64$ & 82.5 & 2.13 & -- & -- \\
$R = 20$ & 83.5 & 2.61 & -- & -- \\
$R = 40$ & 83.4 & 2.58 & -- & -- \\
No $\rho$ refresh & 83.6 & 2.70 & -- & -- \\
\bottomrule
\end{tabular*}
\end{table}

\textbf{Grid keep.} Without grid keep, \benchC loses 2.1 R@10, while \benchA and \benchB change by at most 0.4. Tiles of $3{\times}3$ blocks match or exceed the accuracy of $2{\times}2$ tiles on every history with less computation, and $4{\times}4$ tiles lose 1.4 R@10 on \benchC. With $3{\times}3$ tiles, \method keeps 63\%, 69\%, and 71\% of the blocks on \benchA, \benchB, and \benchC.

\textbf{Pixel tolerances.} The runs of Table~\ref{tab:sens-tol} encode screenshots in fp16, so their reference row differs from the main results by up to 0.3 R@10. Raising both tolerances from 8 to 128 levels lowers computation 1.7$\times$ on \benchB and 2.0$\times$ on \benchC at 1.0 and 2.0 R@10. Over all settings we tried, R@10 changes by at most 1.1 on \benchB and 2.0 on \benchC. Changing $R$ or $\rho$ moves R@10 by at most 0.2 and computation by at most 4\%.

\textbf{Dropping more blocks.} To drop a set fraction of blocks, as in Figure~\ref{fig:frontier}, we choose $\tau$ for each screenshot by binary search. Drop rates of 50\% and 60\% give 81.9 and 81.2 R@10 on \benchA and 55.8 and 52.4 on \benchC, against 82.3 and 57.4 for the fp16 reference.

\section{Redundancy by Source}
\label{app:redundancy}

\begin{table}[t]
\centering
\small
\setlength{\tabcolsep}{3pt}
\caption{\textbf{Redundancy by source.} Share of blocks (\%) that repeat a neighbor (Spatial), or else are unchanged from the frame 5\,s earlier (Temporal), and their sum (Combined), averaged over consecutive frame pairs. \benchC uses 142 of its 787 recordings.}
\label{tab:redundancy}
\begin{tabular*}{\columnwidth}{@{\extracolsep{\fill}}lrccc@{}}
\toprule
Source & Pairs & Spatial & Temporal & Combined \\
\midrule
\multicolumn{5}{@{}l}{\emph{Screen}} \\
\quad \benchA & 64,855 & 46.9 & 25.5 & 72.4 \\
\quad \benchB & 11,845 & 41.4 & 21.9 & 63.3 \\
\quad \benchC & 6,098 & 39.4 & 36.2 & 75.6 \\
\quad WebLINX & 2,975 & 52.4 & 24.5 & 76.9 \\
\quad GUI-World & 194 & 58.2 & 8.4 & 66.6 \\
\midrule
\multicolumn{5}{@{}l}{\emph{Natural video}} \\
\quad Wikimedia Commons & 1,065 & 17.6 & 1.3 & 18.9 \\
\quad YouTube & 470 & 10.1 & 0.6 & 10.7 \\
\quad EPIC-KITCHENS & 300 & 5.6 & 0.1 & 5.7 \\
\bottomrule
\end{tabular*}
\end{table}

Table~\ref{tab:redundancy} gives the redundancy of \S\ref{sec:redundancy-measure} per source, including the three histories of \bench, measured with the same blocks, interval, and tolerance. The public sources are WebLINX~\citep{lu2024weblinx}, GUI-World~\citep{chen2025gui}, and natural video from Wikimedia Commons, YouTube, and EPIC-KITCHENS-100~\citep{damen2022rescaling}.

\section{Qualitative Examples}
\label{app:qualitative}
\definecolor{okGreen}{HTML}{2E8B57}\definecolor{badRed}{HTML}{C8414B}

Table~\ref{tab:qualitative} shows the screenshot that each index ranks first for four single-recording questions, three from \benchA and one from \benchB. All three indexes use the fine-tuned Qwen3-VL-2B, and the text index embeds the Apple Vision OCR text of each screenshot. We chose the questions among those on which the text index and \method rank different recordings first, so they are not a random sample; Figure~\ref{fig:main} gives the accuracy over all questions.

On all four questions, the screenshot index and \method rank a gold screenshot first, and \method indexes a screenshot in 0.66--0.84\,s, against 2.33--2.34\,s for the screenshot index. The text index ranks the gold recording 16th to 752nd and instead ranks first a screenshot of another task, such as setting a desktop background for the profile-photo question or adding an LG television to a Costco cart for the wishlist question.

\begin{table*}[t]
\centering
\footnotesize
\setlength{\fboxsep}{0pt}\setlength{\fboxrule}{1.2pt}
\renewcommand{\arraystretch}{1.1}
\begin{tabular*}{\textwidth}{@{\extracolsep{\fill}}ccc@{}}
\toprule
Text index & Screenshot index & \method \\
\scriptsize 0.62--0.65\,s per screenshot & \scriptsize 2.33--2.34\,s per screenshot & \scriptsize 0.66--0.84\,s per screenshot \\
\midrule
\multicolumn{3}{@{}l}{\textbf{Question.}\ Which smartphone model did I add to my cart?} \\[2pt]
\fcolorbox{badRed}{white}{\includegraphics[width=0.31\textwidth]{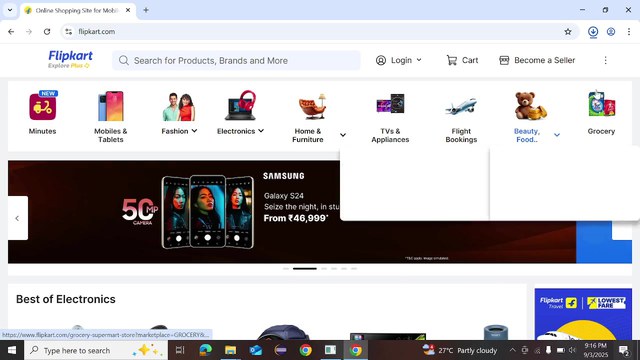}} & \fcolorbox{okGreen}{white}{\includegraphics[width=0.31\textwidth]{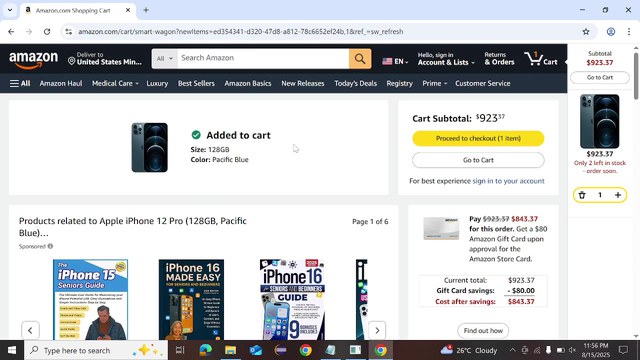}} & \fcolorbox{okGreen}{white}{\includegraphics[width=0.31\textwidth]{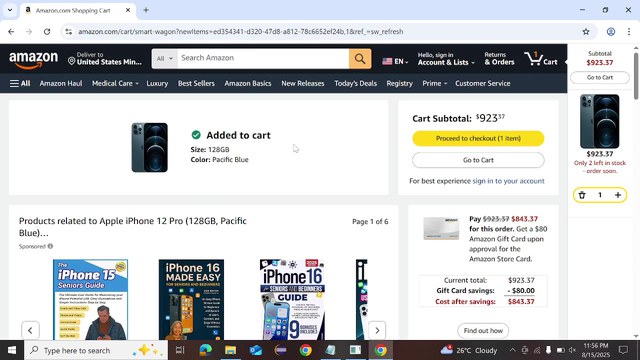}} \\
\midrule
\multicolumn{3}{@{}l}{\textbf{Question.}\ Which work app did I update my profile photo in?} \\[2pt]
\fcolorbox{badRed}{white}{\includegraphics[width=0.31\textwidth]{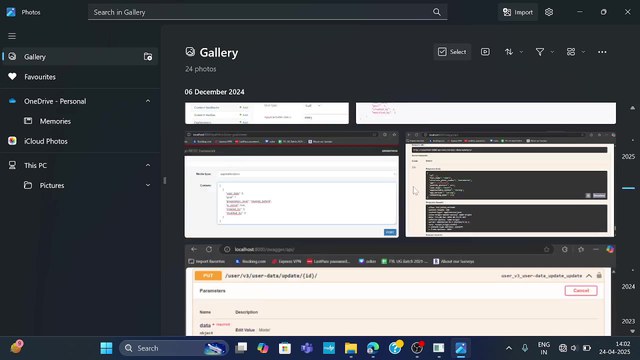}} & \fcolorbox{okGreen}{white}{\includegraphics[width=0.31\textwidth]{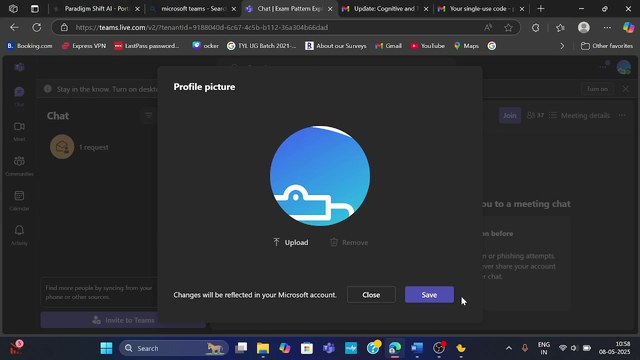}} & \fcolorbox{okGreen}{white}{\includegraphics[width=0.31\textwidth]{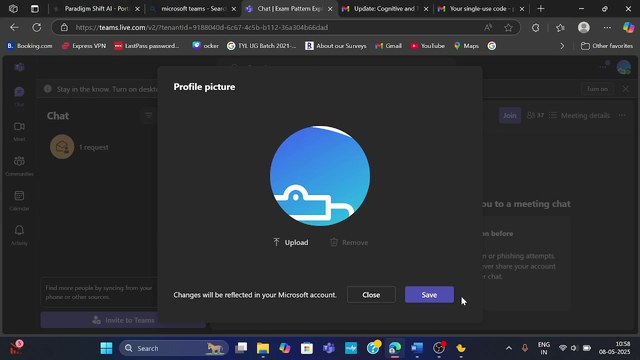}} \\
\midrule
\multicolumn{3}{@{}l}{\textbf{Question.}\ Which all-in-one desktop did I look up the starting price of?} \\[2pt]
\fcolorbox{badRed}{white}{\includegraphics[width=0.31\textwidth]{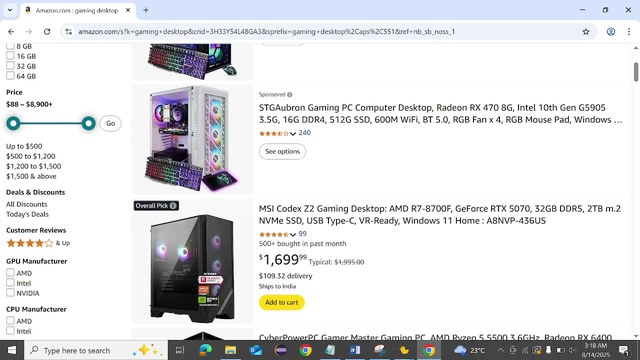}} & \fcolorbox{okGreen}{white}{\includegraphics[width=0.31\textwidth]{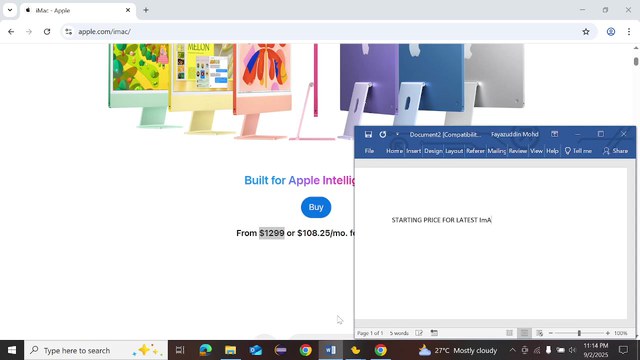}} & \fcolorbox{okGreen}{white}{\includegraphics[width=0.31\textwidth]{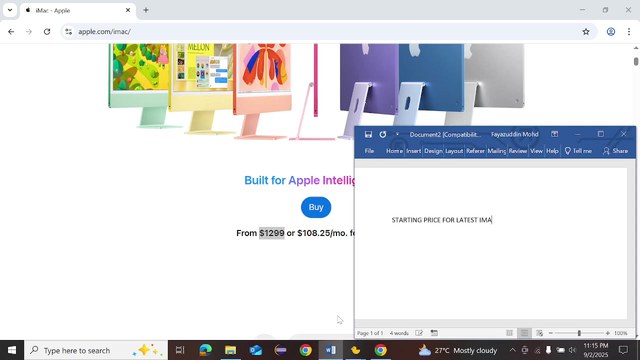}} \\
\midrule
\multicolumn{3}{@{}l}{\textbf{Question.}\ Which television did I add to my newly created wishlist?} \\[2pt]
\fcolorbox{badRed}{white}{\includegraphics[width=0.31\textwidth]{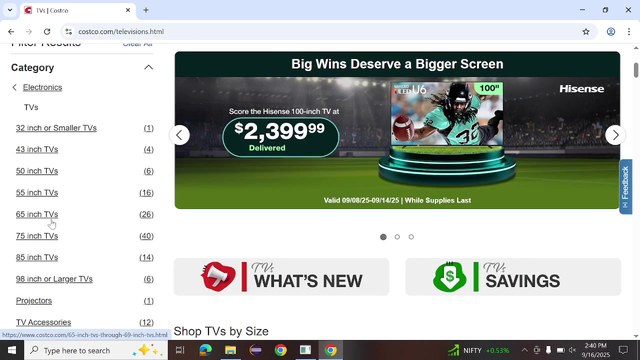}} & \fcolorbox{okGreen}{white}{\includegraphics[width=0.31\textwidth]{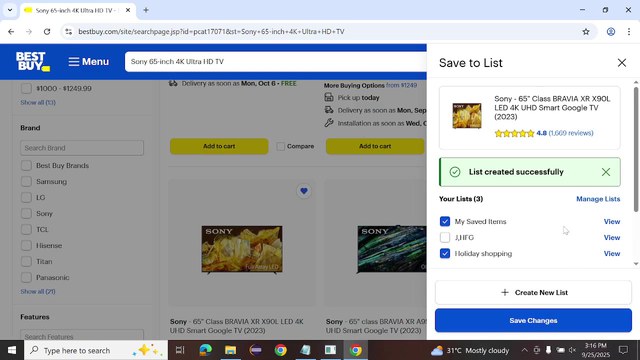}} & \fcolorbox{okGreen}{white}{\includegraphics[width=0.31\textwidth]{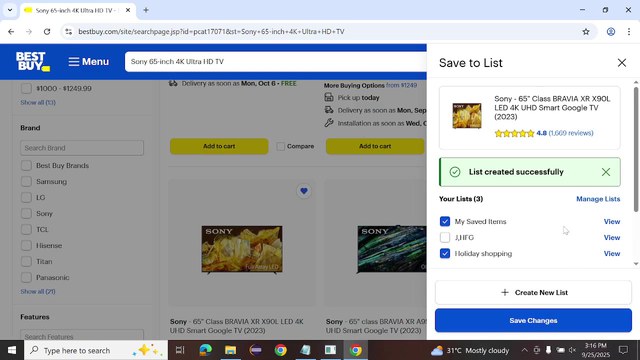}} \\
\bottomrule
\end{tabular*}
\caption{\textbf{Screenshots ranked first for four questions.} For each single-recording question, the screenshot that each index ranks first, framed in green if it belongs to a gold recording and in red otherwise. All indexes use the fine-tuned Qwen3-VL-2B. Below each index is its mean GPU time to index one screenshot on \benchA and \benchB (Figure~\ref{fig:main}).}
\label{tab:qualitative}
\end{table*}

\end{document}